\documentclass[11pt]{article}%

\usepackage{url}
\usepackage{multirow}
\usepackage{array}

\RequirePackage{amsmath,amsfonts, amssymb,amsthm}%
\usepackage{mathtools}%
\usepackage{xcolor}%
\usepackage{mathrsfs} %
\usepackage[a4paper, left=2cm, right=2cm, top=3cm, bottom=3cm]{geometry}%

\usepackage[titletoc,title]{appendix}%
\usepackage[utf8]{inputenc}%
\usepackage{dsfont}%

\usepackage{comment}%
\usepackage{graphicx, color}%

\usepackage[round]{natbib}
\renewcommand{\bibname}{References}

\usepackage{microtype}
\usepackage{graphicx}
\usepackage{subfigure}
\usepackage{booktabs} %

\usepackage{hyperref}
\usepackage{macros}
\usepackage{wrapfig}
\usepackage{caption}
\usepackage[linesnumbered,ruled,vlined]{algorithm2e}
\usepackage{algorithmic}
\usepackage{enumitem}

\usepackage{tikz}
\usetikzlibrary{matrix}
\usepackage{pgfplots}
\usepgfplotslibrary{patchplots}
\usetikzlibrary{patterns, positioning, arrows}
\usepackage{placeins}
\usepackage{todonotes}

\begin{document}

\title{Theoretical Guarantees for Bridging Metric Measure Embedding and Optimal Transport}

\author{
\begin{tabular}{*{4}{>{\centering}p{.45\textwidth}}}
Mokhtar Z. Alaya 								& Maxime Bérar \tabularnewline
LMAC EA 2222 									& LITIS EA 4108 \tabularnewline  
Université de Technologie de Compiègne 		& Université de Rouen Normandie \tabularnewline
\texttt{elmokhtar.alaya@utc.fr} 				& \texttt{maxime.berar@univ-rouen.fr} \tabularnewline
\\
Gilles Gasso  									& Alain Rakotomamonjy \tabularnewline
LITIS EA 4108									& LITIS EA 4108\tabularnewline 
Université de Rouen Normandie\\ \& INSA Rouen 	& Université de Rouen Normandie\\ \& Criteo AI Lab \tabularnewline
\texttt{gilles.gasso@insa-rouen.fr} 			& \texttt{alain.rakoto@insa-rouen.fr} \tabularnewline
\end{tabular}
}

\maketitle

\begin{abstract}
We propose  a novel approach for comparing distributions whose supports
do not necessarily lie on the same metric space. Unlike Gromov-Wasserstein (GW) distance which compares pairwise distances of elements from each distribution, we consider a method allowing to embed the metric measure spaces in a common Euclidean space and compute an optimal transport (OT) on the embedded distributions. This leads to what we call a \emph{sub-embedding robust Wasserstein} (SERW) distance. Under some  conditions,  SERW  is a  distance that considers an OT distance of the (low-distorted) embedded  distributions using a common metric. In addition to this novel proposal that  generalizes several recent
OT works, our contributions stand on several theoretical analyses: {\it (i)} we characterize the embedding spaces to define SERW distance for distribution alignment; {\it (ii)} we prove that SERW mimics almost the same properties of GW distance, and we give a cost relation between GW and SERW. The paper also provides some numerical {illustrations of how SERW behaves on matching problems.}
\end{abstract}

\section{Introduction} %
\label{sec:introduction}

Many central tasks in machine learning often attempt  to align or match real-world entities, based on computing distance (or dissimilarity) between pairs of corresponding probability distributions. Recently, optimal transport (OT) based data analysis has proven a significant usefulness to achieve such tasks, arising from designing loss functions~\citep{frogner2015nips}, unsupervised learning~\citep{pmlr-v70-arjovsky17a}, clustering~\citep{ho2017}, text classification~\citep{kusnerb2015}, domain adaptation~\citep{courty2017optimal}, computer vision~\citep{bonnel2011,solomon2015}, among many more applications~\citep{klouri17,peyre2019COTnowpublisher}. Distances based on OT are referred to as the Monge-Kantorovich or Wasserstein distance~\citep{monge1781,kantorovich1942,villani09optimal}. OT tools allow for a natural geometric comparison of distributions, that takes into account the metric of the underlying space to find the most cost-efficiency way to transport mass from a set of sources to a set of targets. The success of machine learning algorithms based on Wasserstein distance is due to its nice properties~\citep{villani09optimal} and to recent development of efficient computations using entropic regularization~\citep{cuturinips13,genevay2016stochOT,altschulernips17,alaya2019}.

Distribution alignment using Wasserstein distance relies on the assumption that the two sets of entities in question belong to the same ground space, or at least pairwise distance between them can be computed.
To overcome such limitations, one seeks to compute Gromov-Wasserstein (GW) distance~\citep{sturm2006,memoli2011GW}, which is a relaxation of Gromov-Hausdorff distance~\citep{memoli2008,bronstein2010}. GW distance allows to learn an optimal transport-like plan by measuring how the distances between pairs of samples within each ground space are similar. The GW framework has been used for solving alignment problems in several applications, for instance shape matching~\citep{memoli2011GW}, graph partitioning and matching~\citep{xu2019neurips}, matching of vocabulary sets between different languages~\citep{alvarezmelis2018gromov}, generative models~\citep{pmlr-v97-bunne19a}, or matching weighted networks~\citep{chowdhury2018}, to name a few.
{However, computing GW distance induces a heavy computation burden as the underlying problem is a non-convex quadratic program and NP-hard~\citep{peyre2019COTnowpublisher}.}
~\cite{peyre2016Gromov-W} propose an entropic version called {entropic GW discrepancy}, that leads to approximate GW distance.

In this paper, we develop a distance, that is similarly to GW distance applied on sets of entities
from different spaces. Our proposal builds upon metric embedding that allows for an approximation  of some ``hard'' problem with complex metric spaces into another one involving ``simpler'' metric space~\citep{matousek2002} and upon Wasserstein OT cost on the embedding space.  Hence, unlike GW distance that compares pairwise distances of elements from each distribution, we consider a method that embeds the metric measure spaces into a common Euclidean space and computes a Wasserstein OT distance between the embedded distributions. In this context, we introduce a distance, robust to isometry in the embedding space, that  generalizes the ``min-max'' robust OT problem recently introduced in~\cite{patycuturi2019}, where the authors consider
orthogonal projections as embedding functions. 
Main contributions of this work are summarized in the following three points:

\begin{itemize}

	\item We propose a framework  for distribution alignment from different spaces using a sub-embedding robust Wasserstein  (SERW) distance.  As central contribution, we develop the theoretical analysis characterizing the embedding spaces so that SERW be a  distance.

	\item We provide mathematical evidence on the relation
	between GW and our SERW distances. We show for instance, that one key point for approximating GW is that the embeddings be distance-preserving.

	\item We %
	{sketch a potential}
	algorithm describing how our distance can be computed in practice and present numerical %
	{illustrations}
	on simulated and real datasets that support our theoretical results.
\end{itemize}

The remainder of the paper is organized as follows. In Section~\ref{sec:preliminaries} we introduce the definitions of Wasserstein and GW distances, and  set up the embedding spaces. In Section~\ref{sec:proposed_approach} we investigate metric measure embedding for non-aligned distributions through an OT via SERW distance. 
Section~\ref{sec:numerical_experiments} is dedicated to numerical experiments on matching tasks based on simulated and real data. The proofs of the main results are postponed to the appendices.

\section{Preliminaries} %
\label{sec:preliminaries}

We start here by reviewing basic definitions of the materials needed to introduce the main results.
We consider two metric measure spaces (mm-space for short)~\citep{gromov1999metric} $(X, d_X, \mu)$ and $(Y, d_Y, \nu)$, where  $(X, d_X)$ is a compact metric space  and $\mu$ is a probability measure with full support, i.e. $\mu(X) = 1$ and $\text{supp}[\mu] = X$. 
We recall that the support of a measure $\text{supp}[\mu]$ is the minimal closed subset $X_0 \subset X$ such that $\mu(X \backslash X_0) = 0$. Similarly, we define the mm-space $(Y, d_Y, \nu)$. %
Let $\mathscr{P}(X)$ be the set of probability measures in $X$ and $p\in\{1,2\}$. We define $\mathscr{P}_p(X)$ as its subset consisting of measures with finite $p$-moment, i.e.,
\begin{equation*}
	\mathscr{P}_p(X) = \big\{\eta \in \mathscr{P}(X): M_p(\eta) < \infty\big\}, %
\end{equation*}
where $M_p(\eta):=\int_{X} {d^p_X(x, 0)} \diff\eta(x)$. %
For $\mu \in \mathscr{P}(X)$ and $\nu \in \mathscr{P}(Y)$, we write $\Pi(\mu, \nu) \subset \mathscr{P}(X \times Y)$ for the collection of probability measures (couplings) on $X \times Y$ as 
\begin{align*}
	\Pi(\mu, \nu) = \big\{&\pi\in \mathscr{P}(X\times Y):
	 \forall A \subset X, B \subset Y, %
	 \pi(A \times Y) = \mu(A) \text{ and } \pi(X \times B) = \nu(B)\big\}.
\end{align*}

\paragraph{Wasserstein distance.} 

The Monge-Kantorovich or the $2$-Wasserstein distance aims at finding an optimal mass transportation plan $\pi \in \mathscr{P}(X \times Y)$ such that the marginals of $\pi$ are respectively $\mu$ and $\nu$, and these two distributions are supposed to be defined over the {same ground space}, i.e., $X=Y$. It reads as
\begin{equation}
\label{wasserstein-dist}
\mathcal{W}_2^2(\mu, \nu) = \inf_{\pi \in \Pi(\mu, \nu)}\int_{X\times X}d^2_X(x, x')\diff\pi(x,x').
\end{equation}
The infimum in~\eqref{wasserstein-dist} is attained, and any probability $\pi$ which realizes the minimum is called an {\it optimal transport plan.}

\paragraph{Gromov-Wasserstein distance.} 

In contrast to Wasserstein distance, GW one deals with measures that do not necessarily belong to the same ground space.
It learns an optimal transport-like plan which transports samples from a source metric space $X$ into a target metric space $Y$  by measuring how the distances between pairs of samples within each space are similar. 
Following the pioneering work of~\cite{memoli2011GW},  GW distance is defined as 
\begin{align*}
&\mathcal{GW}_2^2(\mu, \nu) = \frac 12 \inf_{\pi \in \Pi(\mu, \nu)} J_\pi(\mu, \nu)%
\end{align*}
where 
\begin{align*}
J_\pi(\mu, \nu) =\iint\limits_{X^2\times Y^2}\ell(d_X(x, x'),d_Y({y,y'}))\diff\pi(x,y) \diff\pi(x',y')
\end{align*}
with a quadratic loss function $\ell(a,b) = |a-b|^2.$ \cite{peyre2016Gromov-W} propose an entropic version called {entropic GW  discrepancy}, allowing to tackle more flexible losses $\ell$, such as mean-square-error or Kullback-Leibler  divergence. The latter version includes an  entropic regularization of $\pi$ in the GW distance computation problem.

\paragraph{Metric embedding.} 

Metric embedding consists in characterizing a new representation of the samples based on the concept of distance preserving.
\begin{definition}
	A mapping $\phi: (X, d_X) \rightarrow (Z, d_Z)$ is an {\it embedding with distortion $\tau$}, denoted as $\tau$-embedding, if the following holds: there exists a constant $\kappa > 0$ (``scaling factor'') such that for all $x, x' \in X,$
\begin{equation}
\label{embedding-ineq}
	\kappa \, d_X(x,x') \leq d_Z(\phi(x), \phi(x')) \leq \tau \kappa \, d_X(x, x').
\end{equation}
\end{definition}
The approximation factor in metric embedding  depends on a distortion parameter of the $\phi$- embedding. This distortion is defined as the infimum of all $\tau \geq 1$ such that the above condition (\ref{embedding-ineq}) holds. If no such $\tau$ exists, then the distortion of $\phi$ is infinity.

In this work we will focus on target spaces $Z$ that are normed spaces endowed with Euclidean distance. Specifically, for some integer $d$ to be specified later, we will consider the metric space $(Z=\R^d, d_Z = \norm{\cdot})$. Hence, one can always take the scaling factor $\kappa$ to be equal to $1$ (by replacing $\phi$ by $\frac 1\kappa\phi$).
Note that an embedding $\phi$ with distortion at most $\tau < \infty$ is necessarily {\it one-to-one} (injective). Isometric embeddings are for instance embeddings with distortion $1.$ For more details about embeddings, we invite the reader to look at the technical report of~\cite{matousek2013}.

We suppose hereafter  $\kappa=1$ in~\eqref{embedding-ineq} and we denote  by $\mathcal{F}_d(X)$ and $\mathcal{F}_d(Y)$ the set of $\tau_\phi$-embedding $\phi:X \rightarrow \R^d$ and $\tau_\psi$-embedding $\psi:Y\rightarrow \R^d$, respectively. We further assume that $\phi(0) = \psi(0) = 0$.
It is worth noting that when $X$ and $Y$ are finite spaces, then the embedding spaces $\mathcal{F}_d(X)$ and $\mathcal{F}_d(Y)$ are non empty. Indeed, suppose we are given a set of $n$ data points $\{x_1, x_2, \ldots, x_n\} = X$, then {\it Bourgain’s embedding theorem}~\citep{bourgain1985} guarantees the existence of an embedding $\phi:X \rightarrow (\R^d, \norm{\cdot})$ with tight distortion at most $\bigO(\log n)$, i.e., $\tau_\phi = \bigO(\log n)$\footnote{There exists an absolute constant $C > 0$ such that $\tau_\phi \leq C \log n.$}, and the target dimension $d = \bigO(\log^2 n)$.
We stress that $d$ is independent of the original dimensions of $X$ and $Y$ and depends only on the number of the given data points $n$ and $m$ and the accuracy-embedding parameters $\tau_\phi$ and $\tau_\psi$. Hence for data points $X = \{x_1, x_2, \ldots, x_n\}$  and $Y = \{y_1, y_2, \ldots, x_m\}$ underlying the distributions of interest, one has 
\begin{equation}
\label{target-dimension}
 d = \bigO(\log^2 (\max(n,m)).
\end{equation}

\begin{remark}
For our setting, we cannot use the Johnson-Lindenstrauss flattening lemma~\citep{Johnson1984ExtensionsOL} (JL-lemma in short) to guarantee the existence of the embeddings. To see that, let us first state this lemma: the JL-lemma  asserts that if $X$ is a { Hilbert space}, $\varepsilon >0$, $n\in \mathbb N$, and $x_1, \ldots x_n \in X$ then there exists a {linear mapping} $\phi_{\text{JL}}: X \rightarrow \R^d$ with $d=\bigO(\log n)$ such that for all $i,{i'} \in \{1, \dots, n\},$ one has 
\begin{equation*}
\|x_i-x_{i'}\|^{2} \leq \|\phi_{\text{JL}}(x_i)-\phi_{\text{JL}}(x_{i'})\|^{2} \leqslant(1+\varepsilon)\|x_i-x_{i'}\|^{2}.
\end{equation*}
Note that the origin space $X$ is supposed to be a Hilbert space in the JL-lemma and the embedding $\phi_{\text{JL}}$ is a linear mapping with a target dimension $d\leq C_\varepsilon \log n$. However, $X$ is only supposed to be a metric space and the target dimension $d\leq C\log^2 n$ in Bourgain's theorem.  Furthermore, the JL-lemma fails to hold true in certain non-Hilbertian settings~\citep{matousek1996, brinkman2005}.
In our approach, we consider a general setup that relies on non-linear embeddings that act on different data structures like text, graphs, images, etc. Therefore, we used Bourgain's theorem to guarantee the non-emptiness of the embedding spaces $\mathcal{F}_d(X)$ and $\mathcal{F}_d(Y)$.
\end{remark}

Let's now highlight all the above criteria characterizing the metric embeddings we consider to define our novel distance and that help us shape some of its properties.
\begin{assumption} 
\label{assumptiom-embeddings}
Assume that $(X,d_X, \mu)$ and $(Y,d_Y,\nu)$ are finite mm-spaces containing the origin $0$, and endowed with measures $\mu$ and $\nu$. %
Assume also that $X$ and $Y$ are of cardinalities $n$ and $m$, the target dimension $d$ satisfies~\eqref{target-dimension}, $\mathcal{F}_d(X) = \{\phi:X \rightarrow \R^d,\tau_\phi\text{-embedding}, \text{ with } \phi(0) = 0\}$ and $\mathcal{F}_d(Y) = \{\psi:Y\rightarrow \R^d, \tau_\psi\text{-embedding}, \text{ with } \psi(0) = 0\}.$
The distortions parameters $\tau_\phi \in \mathcal{D}_{\text{emb}}(X)$, $\tau_\psi \in \mathcal{D}_{\text{emb}}(Y)$ where $\mathcal{D}_{\text{emb}}(X) = [1, \bigO(\log(n))]$ $\text{ and }\mathcal{D}_{\text{emb}}(Y) = [1, \bigO(\log(m))].$
\end{assumption}

\section{Metric measure embedding and OT for distribution alignment} %
\label{sec:proposed_approach}

Let us give first the overall structure of our approach of non-aligned distributions, which generalizes %
recent works~\citep{alvarezmelis2018gromov,patycuturi2019}. 
{
The generalization stands on the fact that the two distributions lie on two different metric spaces and this fact raises several challenging questions about the characterization of the embeddings for yielding a distance. In our approach, we consider a general setup that relies on non-linear embeddings before aligning the measures.
Note, if the metric spaces coincide for both distributions and the embeddings are restricted to be linear (subspace projection) then, our distance reduces to the one proposed by~\cite{patycuturi2019}.
}

In this work, we aim at proposing a novel distance between
two measures defined on different mm-spaces. This distance will be defined as the optimal objective of some optimization problem, we provide technical details and conditions ensuring its existence in the first part of this
section. 
We then present formally our novel distance and its properties including its cost relation with GW distance. 

In a nutshell, our distribution alignment distance between $\mu$ and $\nu$ is obtained as a Wasserstein distance between pushforwards (see Definition~\ref{def:pushforward}) of $\mu$ and $\nu$ w.r.t. some appropriate couple of embeddings $(\phi, \psi) \in \cF_d(X)\times \cF_d(Y)$. Towards this end, we need to exhibit some topological properties of the embeddings spaces, leading at first to the existence of the constructed OT approximate distances. 

\subsection{Topological properties of the embedding spaces} %
\label{sub:topological_properties_of_the_embbedding_space}

We may consider the function $\Gamma_X:  \cF_d(X)\times  \cF_d(X) \rightarrow \R_+$ such that $
\Gamma_X(\phi, \phi') = \sup_{x \in X} \norm{\phi(x) - \phi'(x)}$, for each pair of embeddings $\phi, \phi'\in \cF_d(X)$. 
This function defines a proper metric on the space of embeddings $\cF_d(X)$ and it is referred to as the {\it supremum metric} on $\cF_d(X)$.
Indeed, $\Gamma_X$ satisfies all the conditions that define a general metric.  %
We analogously define the metric $\Gamma_Y$ on $\cF_d(Y)$. 
With the aforementioned preparations, the embeddings spaces satisfy the following topological property.

\begin{proposition}
\label{lem:compact-embedding-sets}
$(\cF_d(X), \Gamma_X)$ and $(\cF_d(Y), \Gamma_Y)$ are both compact metric spaces.
\end{proposition}
Endowing the embedding spaces with the supremum metrics is fruitful, since we get benefits from some existing topological results, based on this functional space metric, to prove the statement in Proposition~\ref{lem:compact-embedding-sets}.  

To let it more readable, the proof of Proposition~\ref{lem:compact-embedding-sets} is divided into 5 steps  summarized as follows: first step is for metric property of $\cF_d(X)$; second one shows completeness of $\cF_d(X)$; third establishes the totally boundedness of $\cF_d(X)$, namely that one can recover this space using balls centered on a { finite} number of embedding points; the last is a conclusion using Arzela-Ascoli's Theorem for characterizing compactness of subsets of functional continuous space, see Appendix~\ref{sub:_lem_compact_embedding_sets_} for all theses details and their proofs. 

We now give a definition of pushforward measures.
\begin{definition}
\label{def:pushforward}
	 (Pushforward measure). Let $(S, \mathscr{S})$ and $(T, \mathscr{T})$ be two measurable spaces,
	 $f: S \rightarrow T$ be a mapping, and $\eta$ be a measure on $S$. The pushforward of $\eta$ by $f$, written $f_{\#}\eta$, is the measure on $T$ defined by $f_{\#}\eta(A) = \eta(f^{-1}(A))$ for $A \in \mathscr{T}$. If $\eta$ is a measure and $f$ is a  measurable function, then $f_{\#}\eta$ is a measure.
\end{definition}

\subsection{Sub-Embedding OT} %
\label{sub:sub_embedding_approximation}

Assume Assumption~\ref{assumptiom-embeddings} holds. Following~\citet{patycuturi2019}, we define an embedding robust version of Wasserstein distance between pushforwards $\phi_{\#}\mu \in\mathscr{P}_p(\R^d)$ and $\psi_{\#}\nu \in \mathscr{P}_p(\R^{d})$ for some appropriate couple of embeddings $(\phi, \psi) \in {\cF}_d(X)\times {\cF}_d(Y)$. We then consider the worst possible OT cost over all possible low-distortion embeddings. The notion of ``robustness" in our distance stands from the fact that we look for this worst embedding.

\begin{definition}
\label{def:ERW}
The $d$-dimensional embedding robust 2-Wasserstein distance (ERW) between $\mu$ and $\nu$ reads as 
\begin{align*}
\mathcal{E}^2_d(\mu, \nu)=  \frac 12 \inf_{r \in \cR_d} \sup_{\phi \in {\cF}_d(X), \psi \in {\cF}_d(Y)}\mathcal{W}_2^2\big(\phi_{\#}\mu,(r\circ\psi)_{\#}\nu\big),
\end{align*}
where $\cR_d$ stands for the set of orthogonal mappings on $\R^d$ and $\circ$ denotes the composition operator between functions. 
\end{definition}

The infimum over the orthogonal mappings on $\R^d$ corresponds to a classical orthogonal procrustes problem~\citep{gower1975generalized,pmlr-v89-grave19a}. It learns the best rotation between the embedded points, allowing to an accurate alignment. The orthogonality constraint ensures that the distances between points are preserved by the transformation.

Note that $\mathcal{E}^2_d(\mu, \nu)$ is finite since the considered embeddings are Lipschitz and both of the distributions $\mu$ and $\nu$ have finite $2$-moment due to Assumption~\ref{assumptiom-embeddings}. Next, using results of pushforward measures, for instance see Lemmas~\ref{lem:integration_pushs} and~\ref{lem:image_admis_couplings} in the Appendices, we explicit ERW in Lemma~\ref{lem:second_def_of_ERW}, whereas Lemmas~\ref{lem:existence_ERW} and~\ref{lem:existence_SERW} establish the existence of embeddings that achieve the suprema defining both ERW and SERW.
\begin{lemma}
\label{lem:second_def_of_ERW}
For any $(\phi, \psi) \in \cF_d(X) \times \cF_d(Y)$ and $r\in \cR_d$, let $J_{\phi, \psi, r,\pi}(\mu, \nu) =  \int_{X\times Y} \norm{\phi(x)- r(\psi(y))}^2 \diff\pi(x,y)$.
One has %
\begin{align*}
\mathcal{E}^2_d(\mu, \nu)= \frac 12 \inf_{r \in \cR_d}\sup_{\phi \in \cF_d(X), \psi \in \cF_d(Y)} \inf_{\pi\in \Pi(\mu,\nu)} J_{\phi, \psi, r,\pi}(\mu, \nu).
\end{align*}
\end{lemma}
By the compactness property of the embedding spaces (see Proposition~\ref{lem:compact-embedding-sets}), the set of optima defining $\mathcal{E}^2_d(\mu, \nu)$ is not empty.
\begin{lemma}
\label{lem:existence_ERW} 
There exist a couple of embeddings $(\phi^*, \psi^*) \in \cF_d(X) \times \cF_d(Y)$ and $r^* \in \cR_d$ such that %
\begin{equation*}
\mathcal{E}^2_d(\mu, \nu) = \frac 12 \mathcal{W}_2^2\big(\phi^*_{\#}\mu,(r^* \circ\psi^*)_{\#}\nu\big).
\end{equation*}
\end{lemma}
Clearly, the quantity $\mathcal{E}^2_d(\mu, \nu)$ is difficult to compute, since an OT is a linear programming problem that requires generally super cubic arithmetic operations~\citep{peyre2019COTnowpublisher}. 
Based on this observation, we focus on the corresponding ``min-max'' problem to define the $d$-dimensional sub-embedding robust 2-Wasserstein distance (SERW). For the sake, we make the next definition.

\begin{definition}
\label{def:SERW}
The $d$-dimensional sub-embedding robust 2-Wasserstein distance (SERW) between $\mu$ and $\nu$ is defined as 
\begin{equation*}
\mathcal{S}^2_d(\mu, \nu)= \frac 12\inf_{\pi\in \Pi(\mu,\nu,\pi)}\inf_{r \in \cR_d}\sup_{\phi \in \cF_d(X), \psi \in \cF_d(Y)} J_{\phi, \psi, r,\pi}(\mu, \nu).
\end{equation*}
\end{definition}
Thanks to the minimax inequality, the following holds.
\begin{lemma}
\label{lem:minimax-ineq}
$\mathcal{E}^2_d(\mu, \nu)  \leq \mathcal{S}^2_d(\mu, \nu).$
\end{lemma}

We emphasize that ERW and SERW quantities play a crucial role in our approach to match distributions
in the common space $\R^d$ regarding pushforwards of the measures $\mu$ and $\nu$  realized by a couple of optimal embeddings and a rotation. Optimal solutions for $\mathcal{S}^2_d(\mu, \nu)$ exist. Namely:
\begin{lemma}
\label{lem:existence_SERW}
There exist a couple of embeddings $(\phi^\star, \psi^\star) \in \cF_d(X) \times \cF_d(Y)$ and $r^\star \in \cR_d$ such that %
\begin{equation*}
\mathcal{S}^2_d(\mu, \nu) = \frac 12 \inf_{\pi\in \Pi(\mu,\nu)} J_{\phi^\star, \psi^\star, r^\star,\pi}(\mu, \nu).
\end{equation*}
\end{lemma}
The proofs of Lemmas~\ref{lem:existence_ERW} and~\ref{lem:existence_SERW} rely on the continuity under integral sign Theorem \citep{schilling_2005}, and the compactness property of the embedding spaces, the orthogonal mappings on $\R^d$ space  and the coupling transport plan $\Pi(\mu, \nu)$, see Appendices~\ref{sub:proof_of_lemma_lem:existence_serw} and~\ref{sub:proof_of_lemma_lem:existence_erw} for more details.

Recall that we are interested in distribution alignment for measures coming from different mm-spaces. One hence expects that SERW  mimics some metric properties of GW distance. %
To proceed in this direction, we first prove that SERW defines a proper metric on the set of all {\it weakly isomorphism} classes of mm-spaces. In our setting the terminology of weakly isomorphism means that there exists a pushforward embedding between mm-spaces. If such a pushforward is $1$-embedding the class is called {\it strongly isomorphism}.

\begin{proposition}
\label{prop:SERW_vanish} 
Let Assumption~\ref{assumptiom-embeddings} holds and assume $X\subseteq \R^{D}$ and $Y\subseteq \R^{D'}$ with $D \neq D'$. Then, 
$\mathcal{S}^2_d(\mu, \nu)=0$ happens if and only if the couple of embeddings $(\phi^\star, \psi^\star)$ and $r^\star \in \cR_d$ optima of $\mathcal{S}^2_d(\mu, \nu)$ verify $\mu = ({\phi^\star}^{-1} \circ r^\star\circ\psi^\star)_{\#} \nu \text{ and } \nu  = ({(r^\star\circ\psi^\star)}^{-1} \circ \phi^\star)_{\#} \mu.$
\end{proposition}

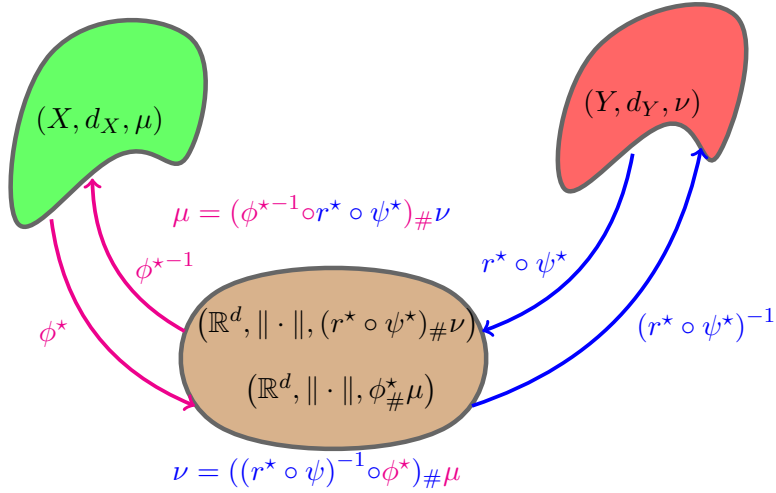
\begin{figure}[htbp]
\centering
\begin{tikzpicture}[scale=0.45]

\draw [smooth cycle, tension=0.8, fill=green,  opacity=0.6, line width=1.7pt]
plot coordinates{(-8,1) (-7,6) (-3,6.5) (-3,3) (-5,3)};
\path[thick, ] (-7,4) [bend left] node[right, xshift=-3.5mm] {${{(X, d_X, \mu)}}$ }(-6,2.2);

\draw [smooth cycle, tension=0.8, fill=red,  opacity=0.6, pattern color=red, line width=1.7pt]
plot coordinates{(8,2) (9,6.2) (13,7.1) (12.5,3) (11,3.9)};
\path[thick, ] (10,4.5) [bend left] node[left, xshift=10.5mm] { ${{(Y, d_Y, \nu)}}$ }(-6,2.2);

\draw[smooth cycle, tension=0.8, fill=brown,  opacity=0.6, pattern color=cyan, line width=1.7pt]
plot coordinates{(-2.1, -1) (-2.1, -5) (4.6, -5) (4.6, -1)};

\path[thick, ] (-2, -2) [bend left] node[right, xshift=-5mm] { {{$\big(\R^d, \norm{\cdot}, (r^\star\circ\psi^\star)_{\#}\nu\big)$}} }(-6,2.2);

\path[thick, ] (-1, -4) [bend left] node[right, xshift=-3mm] { {{$\big(\R^d, \norm{\cdot}, \phi^\star_{\#}\mu\big)$}} }(-6,2.2);

\path[thick, magenta, line width=1.4pt, ->] (-7,1.1) edge [bend right] node[left, xshift=-2mm] { ${{\phi^\star}}$ } (-2.8, -4.4);
\path[thick, magenta, line width=1.4pt, <-] (-5.8,2.3) edge [bend right] node[left, xshift=12mm] { ${{{\phi^\star}^{-1}}}$ } (-3.1, -2.2);

\path[thick, blue, line width=1.4pt, <-] (12, 3.2) edge [bend left] node [right, yshift=-2mm] { ${{{(r^\star\circ\psi^\star)}^{-1}}}$ } (5.3, -4.4);
\path[thick, blue, line width=1.4pt, ->] (9.999, 3.03) edge [bend left] node [right, xshift=-15mm] { ${{{r^\star\circ\psi^\star}}}$ } (5.6, -2.2);

\path[thick, blue, line width=1] (-2, 1.3) [bend left] node[right, xshift=-8mm] {${{\textcolor{magenta}{\mu = ({\phi^\star}^{-1}\circ}\textcolor{blue}{r^\star\circ\psi^\star}\textcolor{magenta}{)_{\#}}\textcolor{blue}{\nu}}}$}(-6,2.2);

\path[thick, magenta, line width=1] (-2, -6.3) [bend left] node[right, xshift=-8mm] {${{\textcolor{blue}{\nu = ({(r^\star\circ\psi)}^{-1}\circ}\phi^\star\textcolor{blue}{)_{\#}}\mu}}$}(-6,2.2);
\end{tikzpicture}
\caption{Illustration of the preserving measure mappings between the mm-spaces $(X, d_X, \mu)$ and $(Y, d_Y, \nu)$  given in Proposition~\ref{prop:SERW_vanish}. The embedding $\phi^\star$ maps from $X$ to $\R^d$ while $r^\star \circ \psi^\star$ maps from $Y$ to $\R^d$. Our distance $\mathcal{S}^2_d(\mu, \nu) =0$ vanishes if and only if $\mu$ and $\nu$ are mapped through the embeddings ${\phi^\star}^{-1}$ and $(r^\star \circ \psi^\star)^{-1}.$}
 
\label{figure_mapping_weak_iso}
\end{figure}

Figure~\ref{figure_mapping_weak_iso} illustrates the mappings between the embedding spaces and how they are assumed to interact in order to satisfy condition in Proposition~\ref{prop:SERW_vanish}.
In~\cite{memoli2011GW} (Theorem 5, property (a)), it is shown that $\mathcal{GW}_2^2(\mu, \nu)=0$ if and only if $(X, d_X, \mu)$ and $(Y, d_Y, \nu)$ are strongly isomorphic. This means that there exists a Borel measurable bijection $\varphi:X \rightarrow Y$  (with Borel measurable inverse $\varphi^{-1}$) such that $\varphi$ is $1$-embedding and $\varphi_{\#}\mu = \nu.$ The statement in Proposition~\ref{prop:SERW_vanish} is a weak version of the aforementioned result, because neither ${\phi^\star}^{-1}\circ r^\star\circ \psi^\star$ nor ${(r^\star \circ\psi^\star)}^{-1} \circ \phi^\star$ are isometric embeddings. However, we succeed to find a measure-preserving mapping relating $\mu$ and $\nu$ to each other via the given pushforwards in Proposition~\ref{prop:SERW_vanish}. 
Note that $r^\star \circ \psi^\star$ maps from the $Y$ space to $\R^d$ while $\phi^\star$ maps from $X$ to $\R^d$. Our distance $\mathcal{S}^2_d(\mu, \nu)$ vanishes if and only if $\mu$ and $\nu$ are mapped through the embeddings ${\phi^\star}^{-1}$ and ${(r^\star\circ\psi^\star)}^{-1}$. %
With these elements, we can now prove that both ERW and SERW are further distances.
\begin{proposition}
\label{prop:ERW_SERW_distances}
Assume that statement of Proposition~\ref{prop:SERW_vanish} holds. Then, ERW and SERW define a proper distances between weakly isomorphism mm-spaces.
\end{proposition}

\subsection{Cost relation between GW and SERW} %
\label{sub:metric_equivalence}

In addition to the afore-mentioned theoretical properties of SERW, we establish a {\it cost relation} metric between GW and SERW distances. The obtained upper and lower bounds depend on approximation constants that are linked to the distortions of the embeddings. 

\begin{proposition}
\label{prop:left_equivalence_GW_SERW} %
Let Assumption~\ref{assumptiom-embeddings} holds. Then, %
\begin{align*}
\frac 12 \mathcal{GW}_2^2(\mu, \nu) \leq \mathcal{S}^2_d(\mu, \nu)+ \alpha\Loverline{M}_{\mu, \nu}
\end{align*}
where $\alpha = 2 \inf_{\tau_{\phi}\in \mathcal{D}_{\text{emb}}(X), \tau_{\psi} \in \mathcal{D}_{\text{emb}}(Y)}(\tau_{\phi}\tau_{\psi}- 1)$ and $\Loverline{M}_{\mu, \nu} = 2(M_1(\mu) + M_1(\nu)).$
\end{proposition}

\begin{proposition}
\label{prop:right_equivalence_GW_SERW}
Let Assumption~\ref{assumptiom-embeddings} holds. Then, %
\begin{align*}
\mathcal{S}^2_d(\mu, \nu) \leq  \beta\mathcal{GW}^2_2(\mu, \nu) + 4\beta \Uoverline{M}_{\mu, \nu}%
\end{align*} 
 where $\beta = 2\sup_{\tau_{\phi}\in \mathcal{D}_{\text{emb}}(X), \tau_{\psi} \in \mathcal{D}_{\text{emb}}(Y)}(\tau_\phi^2 +\tau_\psi^2)$ and 
$\Uoverline{M}_{\mu, \nu} = (\sqrt{M_2(\mu)} + \sqrt{M_1(\mu)}) (\sqrt{M_2(\nu)} + \sqrt{M_1(\nu)}) + M_2(\mu) + M_2(\nu).$ 
\end{proposition}

Proofs of Propositions~\ref{prop:left_equivalence_GW_SERW} and~\ref{prop:right_equivalence_GW_SERW} are presented in Appendices~\ref{sub:proof_of_proposition_prop:right_equivalence_gw_serw} and~\ref{sub:proof_of_proposition_prop:equivalence_gw_serw}. We use upper and lower bounds of GW distance as provided in~\cite{memoli2008}.
The cost relation between SERW and GW distances obtained in Propositions~\ref{prop:left_equivalence_GW_SERW} and~\ref{prop:right_equivalence_GW_SERW} %
are up to the constants $\alpha, \beta$ which are depending on the distortion parameters of the embeddings, and up to an additive constant through the $p$-moments $M_p$ of the measures $\mu$ and $\nu.$ In the following we highlight some particular cases leading to closed form of the upper and lower bounds for the cost relation  between GW and SERW distances.

\subsection{Fixed sub-embedding for distribution alignment}

From the computational point of view, computing SERW distance seems a daunting task, since one would have to optimize over the product of two huge embedding spaces $\cF_d(X)\times \cF_d(Y)$. However in some applications we may not require solving over $\cF_d(X)\times \cF_d(Y)$  and rather have at disposal known embeddings in advance. For instance, for image-text alignment we may leverage on features extracted from pre-trained deep architectures ({VGG} \citep{Simonyan15} for image embedding,  and {Word2vec} \cite{mikolov2013distributed} for projection the text).
Roughly speaking, our SERW procedure with respect to these fixed embeddings can be viewed as an \textit{embedding-dependent distribution alignment} for matching. More precisely, the alignment quality  is strongly dependent on the given embeddings; the lower distorted embeddings, the more accurate alignment.

\begin{definition} %
For a fixed couple of embeddings $(\phi_f, \psi_f) \in \cF_d(X) \times \cF_d(Y)$, we define the fixed sub-embedding robust Wasserstein (FSERW) as
\begin{equation*}
\widetilde{\cS^2_d} = \frac 12\inf_{\pi\in \Pi(\mu,\nu)} \inf_{r \in \cR_d} J_{\phi_f, \psi_f, r, \pi}(\mu, \nu).
\end{equation*}
\end{definition}

\begin{lemma}
\label{lem:fix-emb-distance}
$\widetilde{\cS^2_d}$ defines a proper distance if and only if $\mu = ({\phi_f}^{-1} \circ (r_f\circ\psi_f))_{\#} \nu \text{ and } \nu  = ({(r_f\circ\psi_f)}^{-1} \circ \phi_f)_{\#} \mu,$ where $r_f = \inf_{r\in\cR_d}J_{\phi_f, \psi_f, r, \pi}(\mu, \nu).$
\end{lemma}
The cost relation guarantees given in Propositions~\ref{prop:ERW_SERW_distances} and~\ref{prop:left_equivalence_GW_SERW} are dependent on the distortions of the fixed embeddings, i.e., 
the constants $\alpha$ and $\beta$ become: $\alpha_f = 2(\tau_{\phi_f}\tau_{\psi_f}-1)$ and $\beta_f= 2(\tau_{\phi_f}^2 + \tau_{\psi_f}^2).$ Then the following holds
\begin{lemma}
One has $\frac 12 \mathcal{GW}_2^2(\mu, \nu) \leq \mathcal{S}^2_d(\mu, \nu)+ \alpha_f\Loverline{M}_{\mu, \nu}$ and $\mathcal{S}^2_d(\mu, \nu) \leq  \beta_f\mathcal{GW}^2_2(\mu, \nu) + 4\beta_f \Uoverline{M}_{\mu, \nu}.$
\end{lemma}

In a particular case of isometric embeddings, our procedure gives the following cost relation 
\begin{equation*}
\frac 12 \mathcal{GW}_2^2(\mu, \nu) \leq \mathcal{S}^2_d(\mu, \nu) \leq 4 \mathcal{GW}_2^2(\mu, \nu) 	 + 16\Uoverline{M}_{\mu, \nu}.
\end{equation*}
{%
The additive constants $\Loverline{M}_{\mu, \nu}$ and $\Uoverline{M}_{\mu, \nu}$ can be upper bounded in a setting of data preprocessing, for instance in the case of a $\ell_2$-normalization preprocessing we have $\Loverline{M}_{\mu, \nu} \leq 4$ and $\Uoverline{M}_{\mu, \nu} \leq 6.$}

\section{Numerical experiments} %
\label{sec:numerical_experiments}

Here we illustrate how
SERW distance behaves on numerical problems. We apply it on  some toy problems as well as on some  problems usually addressed using GW distance.

\paragraph{{Sketch} of practical implementation.}
Based on the above presented theory, we have several options for computing the distance between non-aligned measures
and they all come with some guarantees compared to a GW distance.  
In the simpler case of fixed embedding, 
if the original spaces are subspaces of $\R^d$, any distance
preserving  embedding can be a good option for having
an embedding with low distortion. Typically, methods like
multidimensional scaling (MDS)~\citep{kruskal}, Isomap~\citep{tenenbaum2000global} or Local linear embedding (LLE)~\citep{roweis2000nonlinear} can be  good candidates.
One of the key advantages of SERW is that it considers non-linear embedding before measure alignment. Hence, it has  the ability of leveraging over the large zoo
of recent embedding methods that act on different data structures
like text~\citep{grave2018learning}, graphs~\citep{grover2016node2vec,narayanan2017graph2vec}, images \citep{Simonyan15}, or even histograms~\citep{courty2018learning}.

In the general setting our theoretical results require  computing $\mathcal{S}^2_d(\mu, \nu)$ to solve
the problem given in the Definition \eqref{def:SERW}. 
We sketch in Section~\ref{sec:detailed_implementation_and_additional_experiments} a practical procedure to learn from samples $X = \{x_i\}_{i=1}^n$ and $Y = \{y_j\}_{j=1}^m$, non-linear neural-network based embedding functions $\phi$ and $\psi$ that maximize the Wasserstein distance between the embedded samples while minimizing the embedding distortion.

\subsection{Implementation details on learning the embeddings} %
\label{sec:detailed_implementation_and_additional_experiments}

In practice, for computing $\mathcal{S}^2_d(\mu, \nu)$, we need to solve
the problem given in Equation \eqref{def:SERW}. As stated above in some practical situations, we leverage on existing embeddings and consider
the problem without the maximization over the embedings as the space
is restricted to an unique singleton. %
In some other cases, it is possible to learn the embedding
that maximizes the Wasserstein distance between embedded examples and
that minimizes the distance  distortion of the embedding. In what follows,
we detail how we have numerically implemented the computation of
$\mathcal{S}^2_d(\mu, \nu)$ from samples $\{x_i\}$ and $\{y_j\}$ respectively
sampled from $X$ and $Y$ according to $\mu$ and $\nu$.

For all $i,i' \in \{1, \ldots, n\}, j,j' \in \{1, \ldots, m\}$, we denote the pairwise distances $C_{ii'}^X = d_X(x_i, x_{i'}), C_{jj'}^Y = d_Y(y_j, y_{j'})$ in the origin space $X, Y$ and $C_{ii'}^\phi =\|\phi_{\text{}}(x_i)-\phi_{\text{}}(x_{i'})\|^{2}, C_{jj'}^\psi =\|\psi_{\text{}}(y_j)-\psi_{\text{}}(y_{j'})\|^{2}$ in the embbed space $\R^d.$ 
The problem we want
to solve is
\begin{equation}\label{eq:serwnum}
	 \min_{\pi \in \Pi(\mu, \nu)} \min_{r \in \mathcal{R}_d}\max_{\phi,\psi}\Big\{
	\frac{1}{2}\sum_{i,j} \|\phi(x_i) - r(\psi(y_j)) \|^2 \pi_{i,j}
	- \sum_{i,i, i\neq i'} D_X^\phi(C_{i,j}^X,C_{i,j}^\phi) - \sum_{j,j',  j\neq j'} D_Y^\psi(C_{i,j}^Y ,C_{i,j}^\psi) \Big\}
\end{equation}  
with $
\Pi(\mu, \nu)= \{\pi \in \R^{n \times m}| \pi \mathbf 1_m = \mu, \pi^\top \mathbf 1_n = \nu\}$. In this equation, the first sum corresponds to the optimal transport cost function and the other two sums compute the distortion between pairwise distances in the input space and embedded space respectively for the $x$ and the $y$ samples. In the notation, $D_X^\phi(\cdot,\cdot)$ is  a loss function  that 
penalizes the discrepancy between the input $C_{ii'}^X$ and embedded $C_{ii'}^\phi$ distances. This distance loss $D_X^\phi$ has been designed so as to encourage the embedding to preserve pairwise distance up to a 
$\tilde{\tau}_\phi$ factor. Hence 
\begin{equation*}
D_X^\phi(C_{ii'}^X,C_{ii'}^\phi) =  {\ind{}}\Big({\frac{C_{ii'}^\phi}{ C_{ii'}^X} > \tilde{\tau}_\phi}\Big) \frac{C_{ii'}^\phi}{C_{ii'}^X} 
\quad \quad%
D_Y^\psi(C_{jj'}^Y,C_{jj'}^\psi) =  {\ind{}}\Big({\frac{C_{jj'}^\psi}{ C_{jj'}^Y} > \tilde{\tau}_\psi}\Big) \frac{C_{jj'}^\phi}{C_{jj'}^Y} 
\end{equation*}
with 
\begin{equation*}
\min_{i \neq i'}{\frac{C_{ii'}^\phi}{ C_{ii'}^X}} \leq \tilde{\tau}_\phi \leq \max_{i \neq i'}{\frac{C_{ii'}^\phi}{ C_{ii'}^X}}
\quad \quad%
\min_{j \neq j'}{\frac{C_{jj'}^\phi}{ C_{jj'}^Y}} \leq \tilde{\tau}_\psi \leq \max_{j \neq j'}{\frac{C_{jj'}^\psi}{ C_{jj'}^Y}}
\end{equation*}
and ${\ind{} (\cdot)}$ denotes the indicator function.
In the experiments $\tilde{\tau}_\phi$ is fixed as  $\max(0.9 \max_{i \neq i'}{\frac{C_{ii'}^\phi}{ C_{ii'}^X}},\min_{i \neq i'}{\frac{C_{ii'}^\phi}{ C_{ii'}^X}})$ (similarly for $\tilde{\tau}_\psi$). 
It penalizes the embbeded couples of inputs whose embbeded pairwise distances are the most dissimilar to the input pairwise distances. As these specific discrepancies impact the estimation of the distorsion rate of the embedding, the designed loss has been tailored to reduce the distorsion rate comparatively to those of the initial embeddings.

 In practice, the embedding functions $\phi$ and $\psi$ have been implemented in the following way 

\begin{equation} \label{eq:embedding def}
\phi = (\mathbf{I} + g_X)\,\circ\, h_{\theta_X} \quad \quad  \psi = (\mathbf{I} + g_{Y})\,\circ\, h_{\theta_Y} \quad \quad
\end{equation}
where $\mathbf{I}$ is the identity matrix, $g_X : \mathbb{R}^d \rightarrow \mathbb{R}^d$ and  $g_Y : \mathbb{R}^d  \rightarrow \mathbb{R}^d $    are trainable neural networks based embeddings and   $h_{\theta_X}
:  X \rightarrow \mathbb{R}^d $
and $h_{\theta_Y} : Y \rightarrow \mathbb{R}^d $ are
data-dependent low-dimensional projections that preserves
(local) distances. Typically, for the $h$ functions, we have considered in our experiments algorithms like MDS or LLE.
Our choice of the embedding dimension can be then principally giving by the $d$-largest number of eigenvalues of the spectral decomposition arising within these algorithms.

So the learning problem described in Equation \eqref{eq:serwnum} involves a  
max-min problem over the Wasserstein distance of the mapped samples.
For solving the problem, we have adopted an alternate optimization strategy
where for each mini-batch of samples from $\{x_i\}$ and $\{y_j\}$, we first optimize $r$ and $\pi$ at fixed $\phi$ and $\psi$ and then optimize the embeddings for fixed 
optimal $r$ and $\pi$. In practice, the sub-problem with respects to $r$ and $\pi$
is an invariant OT problem and can be solved using the algorithm proposed by \citet{pmlr-v89-alvarez-melis19a}. $g_{X}$ and $g_{Y}$ is implemented
as two fully connected neural networks with leaky ReLU activation functions and no bias. They are optimized using stochastic gradient descent using Adam as optimizer. Some details of the algorithms is provided in Algorithm~\ref{algo:computeSERW}.
\begin{algorithm}[htbp]
	\caption{\textsc{Computing SERW with trained $\phi$ and $\psi$}}
	\label{algo:computeSERW}
	\begin{algorithmic}
		\STATE {\bfseries Input:} Source and target samples: $(X, \mu)$ and $(Y, \nu),$ Epochs $(E)$, batch sizes $(B);$\\
		\STATE {\bfseries Output:} the embeddings $\phi = (\mathbf{I} + g_X)\,\circ\, h_{\theta_X}$ and 
		$\psi = (\mathbf{I} + g_Y)\,\circ\, h_{\theta_Y};$
		\FOR{$e=$1 \textbf{to} $E$}
		\FOR  {{$b=1$} \textbf{to} $B$ }
		\STATE sample batches $x$ and $y$ from the two input spaces;
		\STATE get $x_{\text{emb}} \leftarrow \phi(x)$ and $y_{\text{emb}} \leftarrow \psi(y);$ 
		\STATE get $r^\star$ and $\pi^\star$ by minimizing Equation \eqref{def:SERW} with using $x_{\text{emb}}$ and $y_{\text{emb}};$
		\STATE update $\theta_X$ and $\theta_Y$  by maximizing Equation \eqref{eq:serwnum} using $r^\star$ and $\pi^\star;$ 
		\ENDFOR
		\ENDFOR
		\STATE {\bfseries Return:} {$\pi^{(e)};$}
	\end{algorithmic}
\end{algorithm}

\subsection{Toy example}
\label{sub:toy}

\begin{figure}[htbp]
	\centering
	\includegraphics[width=0.8\textwidth]{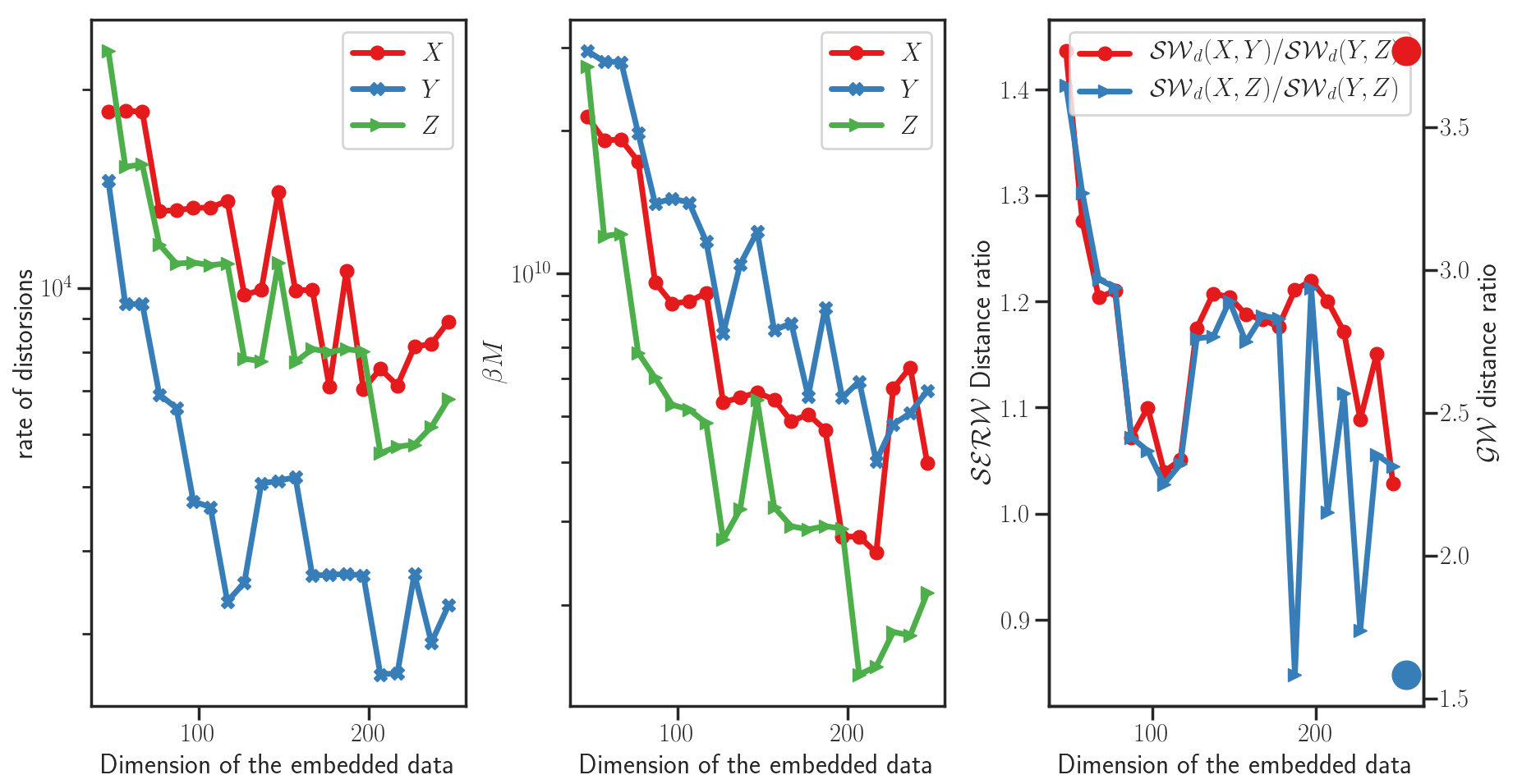}
	\caption{Plots of the distortion rates (left); various bounds in Proposition \ref{prop:right_equivalence_GW_SERW} (middle);  GW cost and the distance ratio of SERW between the data points (right); as a function of the target dimension of embedded data $d$. The bold points in the right panel correspond to $\mathcal{GW}_2(X, Y) / \mathcal{GW}_2(Y, Z)$ (red) and  $\mathcal{GW}_2(X, Z) / \mathcal{GW}_2(Y, Z)$ (blue). Note that the distance ratio (red curve) is upper than the blue one which highlights that the proximity preservation for SERW distance is fulfilled as the embbed dimension $d$ goes up.}
	\label{fig:distorsion}
\end{figure}

In this example, we extracted randomly $n=m=1000$ samples from MNIST, USPS and Fashion MNIST data sets, denoted by ${X}$, ${Y}$ and ${Z}$. We compare GW distances between three possible matchings with the assorted SERW distances. We pre-process the data in order to fix the parameter $\Loverline{M}_{\mu, \nu}$ and $\Uoverline{M}_{\mu, \nu}$ as discussed previously. We then vary the dimension of the embedded points from $\log^2 n$ up to the smallest dimension of the original samples. We perform the embeddings by using LLE followed by a non-linear embedding scheme aiming at minimizing the distance distortion as described in Section~\ref{sec:detailed_implementation_and_additional_experiments}.

In Figure~\ref{fig:distorsion} we report plots of the distortion rate, the additive  constant $\beta \Uoverline{M}_{\mu, \nu}$ in the upper bound in Proposition~\ref{prop:right_equivalence_GW_SERW}, and the distance ratio of SERW for the three data sets $X$, $Y$ and $Z$. As can be seen the rates decrease as the embedding dimension increase. Note that to determine the distortion coefficient for each given embedded dimension, we  compute the quotient of the pairwise distances both in the original and embedding spaces. Thus, this high magnitudes of the upper bounds are due to a ``crude'' estimation of the distortion rate. One may explore a better estimation to reach a tighter upper bound. For this toy set, we investigate a useful property in our approach called {\it proximity preservation}, a property stating that: 
\begin{align*}
\mathcal{GW}_2(\mu, \nu) \leq \mathcal{GW}_2(\mu, \eta) \Rightarrow \mathcal{S}_d(\mu, \nu) \leq \mathcal{S}_d(\mu, \eta).
\end{align*}
In order to confirm this property, we compute the ratio between $\mathcal{S}_d(X,Y)/\mathcal{S}_d(Y,Z)$ and $\mathcal{S}_d(X,Z)/\mathcal{S}_d(Y,Z)$ for various embeddings and compare the resulting order with the quantities $\mathcal{GW}_2(X,Y)/\mathcal{GW}_2(Y,Z)$ and $\mathcal{GW}_2(X,Z)/\mathcal{GW}_2(Y,Z)$. As seen in Figure \ref{fig:distorsion}, while the ratios vary their order is often preserved for large embedding dimensions.

\subsection{Meshes comparison} 
\label{sub:mesh}

\begin{figure}[htbp]
	\centering
		\includegraphics[width=0.6\textwidth]{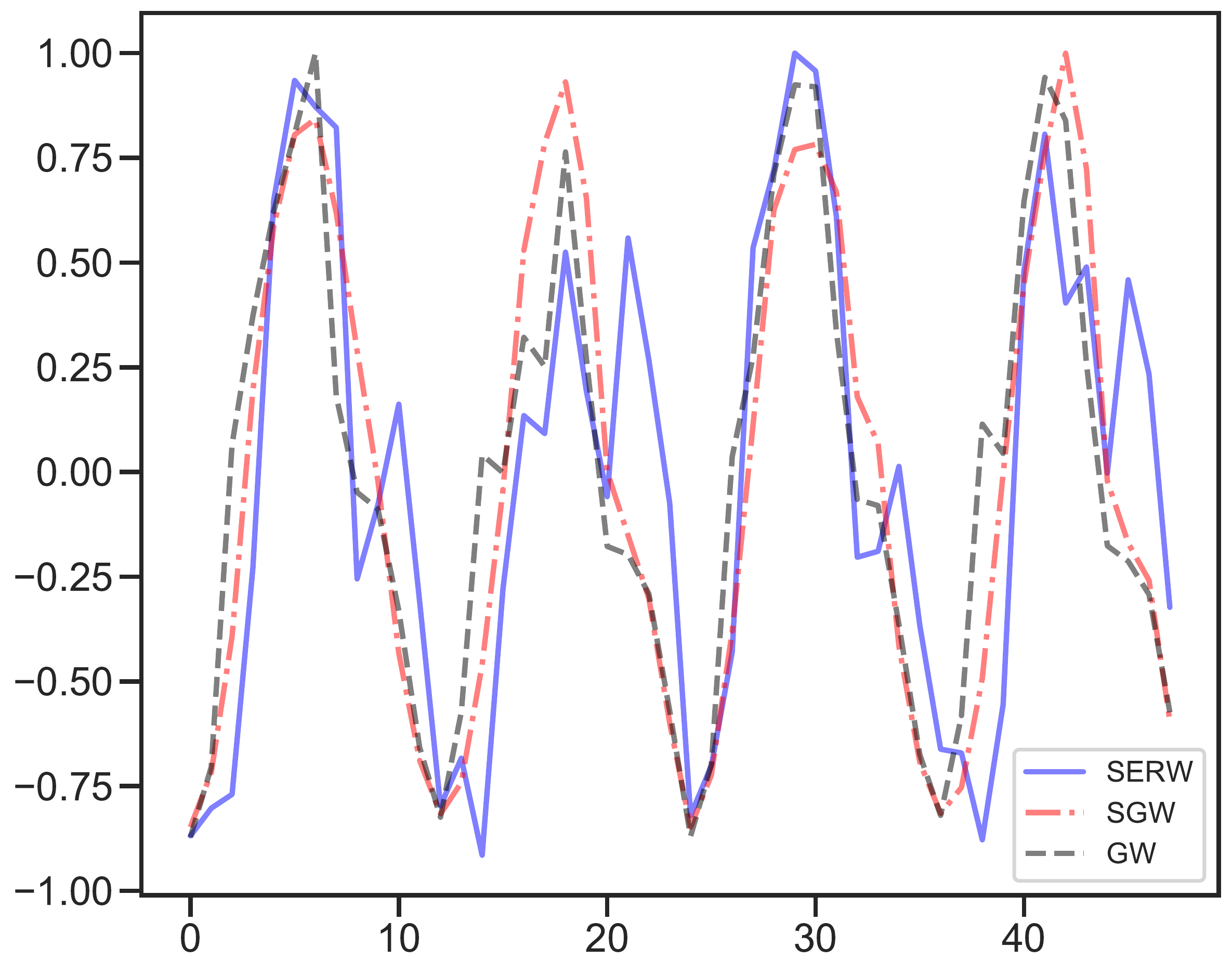}
	\caption{GW, SGW, and SERW distances between 3D meshes of galloping horses. 
	We can note that both SGW and SERW distances are able to retrieve the cyclic nature of the movement (see also Figure~\ref{fig:dtw-distances}).   
		\label{fig:meshes}}
\end{figure}

GW distance is frequently used in computer graphics for 
computing correspondence between meshes. Those distances are then exploited for
organizing shape collection, for instance for shape retrieval or search. One of
the useful key property of GW distance for those applications is that it
is isometry-invariant. In order to show that our proposed approach approximately
satisfies this property, we reproduce an experiment already considered by~\cite{solomon2016entropic}
and~\cite{sgw}.

We have at our disposal  a time-series of $45$ meshes of galloping horses. When
comparing all meshes with the first one, the goal is to show that the distance presents
a cyclic nature related to galop cycle.  Each mesh is composed of $8400$ samples and
in our case, we have embedded them  into a $2$-dimensional space using a multi-dimensional scaling algorithm followed by a non-linear embedding aiming at minimizing distortion as described in  Section~\ref{sec:detailed_implementation_and_additional_experiments}.   

Figure \ref{fig:meshes} shows the (centered and max-normalized) distances between meshes we obtain with {SERW} and with a Sliced Gromov-Wasserstein (SGW)~\citep{sgw}  and a genuine GW distance. In both cases, due to the random aspect of the first two algorithms, distances are averaged over $10$ runs. We note that our approach is able to recover the cyclic nature
of the galloping motion as described by GW distance.  

\begin{figure}[htbp]
	\centering
	\includegraphics[width=0.45\textwidth]{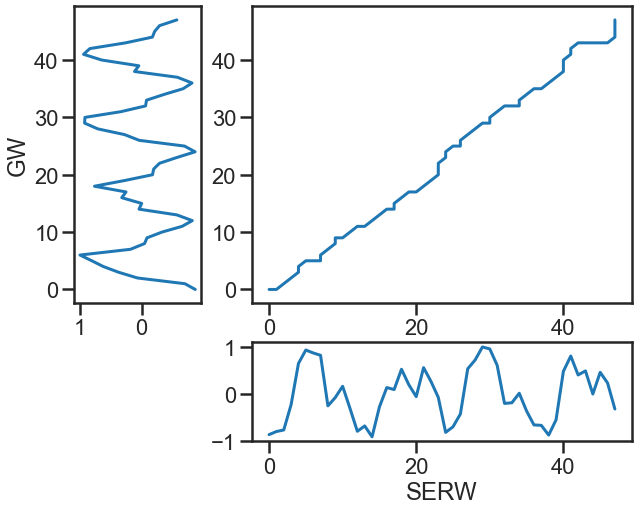}
	\includegraphics[width=0.45\textwidth]{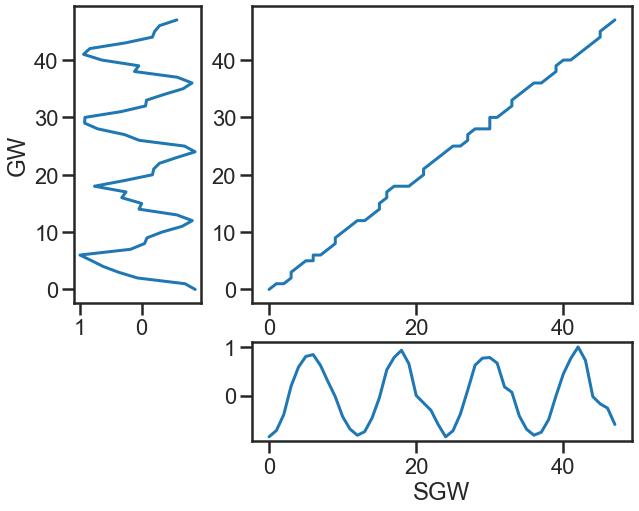}
	\caption{Alignment of  the couple distances (GW, SERW) and (GW, SGW) between 3D meshes images as yielded by DTW measure. Optimal alignment is diagonal. On the right and bottom of each panel are the compared curves.}
	\label{fig:dtw-distances}
\end{figure}

To quantify the cyclic nature
of the galloping motion as described by GW, SERW and SGW, we compute the dynamic time warping measures (DTW)~\citep{berndtDTW1994}. DTW aims to find an optimal alignment between two given signals. Figure~\ref{fig:dtw-distances} illustrates that the couple distances (GW, SERW) and (GW, SGW) behave well according to the recovery of galloping motion. Indeed, the alignment suggested by DTW in both cases is almost diagonal showing that the motion curves provided by SERW and SGW are in accordance with the obtained motion curve via the plain GW. We further compute the mean and the standard deviation of DTW measures between GW and SERW one the one hand, and between GW and SGW distances on the other hand, over the $10$ runs. We get the following: $\text{DTW(GW, SERW)} = 10.62 \pm 1.35$ and $\text{DTW(GW, SGW)} = 10.57 \pm 1.89$. This emphasizes the fact that SERW retrieves the cyclic nature of the movement as does by GW.

\subsection{Text-Image alignment}

To show that our proposed also provides relevant coupling when considering out-of-the-shelves embeddings, we present here results on aligning text and images distributions. The problem we address is related to identifying different states of objects, scene and materials \citep{StatesAndTransformations}.  %
We have images labeled by some nouns modified by some adjectives describing 
state of the objects.  In our experiment, we want to show that our
approach provides coupling between  labels and images semantically meaningful as %
those obtained by a Gromov-Wasserstein approach.  As for proof of concept, 
from the $115$ available adjectives, we have considered only three of them 
\emph{ruffled, weathered, engraved} and extracted all the classes associated
with those adjectives. In total, we obtain $109$ different classes of objects and about $525$ images in total (as each class contains at most $5$ objects).

The composed name (adjective + noun) of each label is embedded into $\mathbb{R}^{100}$ using a word vector representation issued by \texttt{fasttext} model~\citep{grave2018learning} trained on the first 1 billion bytes of English Wikipedia according to~\cite{mikolov2018advances}. The $256 \times 256$ images have been embedded into a vector of dimension $4096$ using a pre-trained VGG-16 model~\citep{Simonyan15}. These embeddings are extracted  from the first dense layer of a VGG-16. 
\begin{figure}[htbp]
\centering
\includegraphics[width=0.65\textwidth]{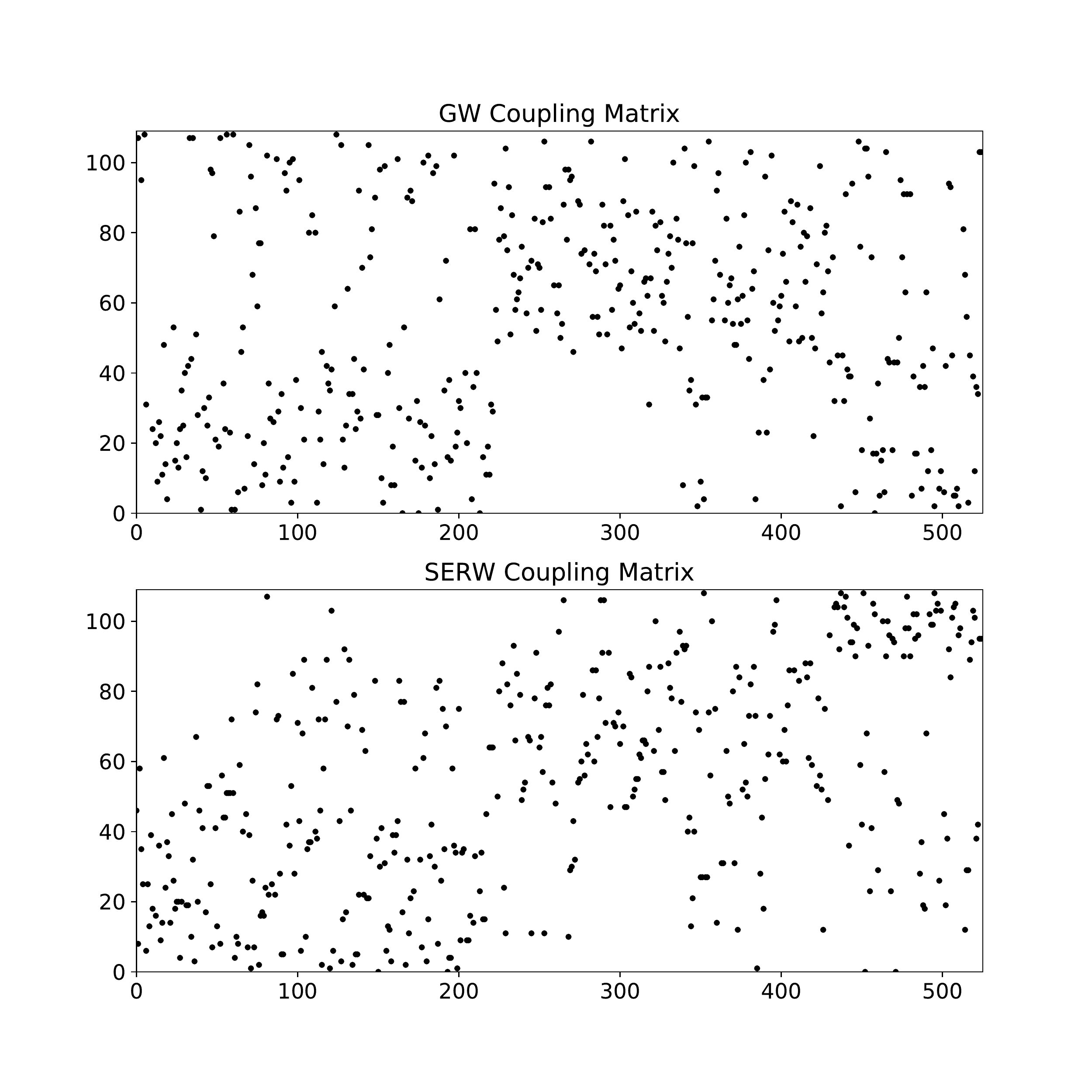} %
\caption{Coupling matrices between text and image embeddings. (top) Gromov-Wasserstein coupling matrix obtained in the original embedding spaces (bottom) our SERW coupling matrix after projecting embeddings into same dimension space.\label{fig:gwtextimage}}
\end{figure}
The Gromov-Wasserstein distance of those embeddings has been computed for coupling labels and images in the two different embedding spaces. For our SERW approach, we have further reduce the dimension of the image embeddings using  Isomap with $100$ dimensions. When computing the distance matrix, objects have been organized by class of adjectives for an easy visual inspection.

\begin{figure}[htbp]
	\includegraphics[width=0.95\linewidth]{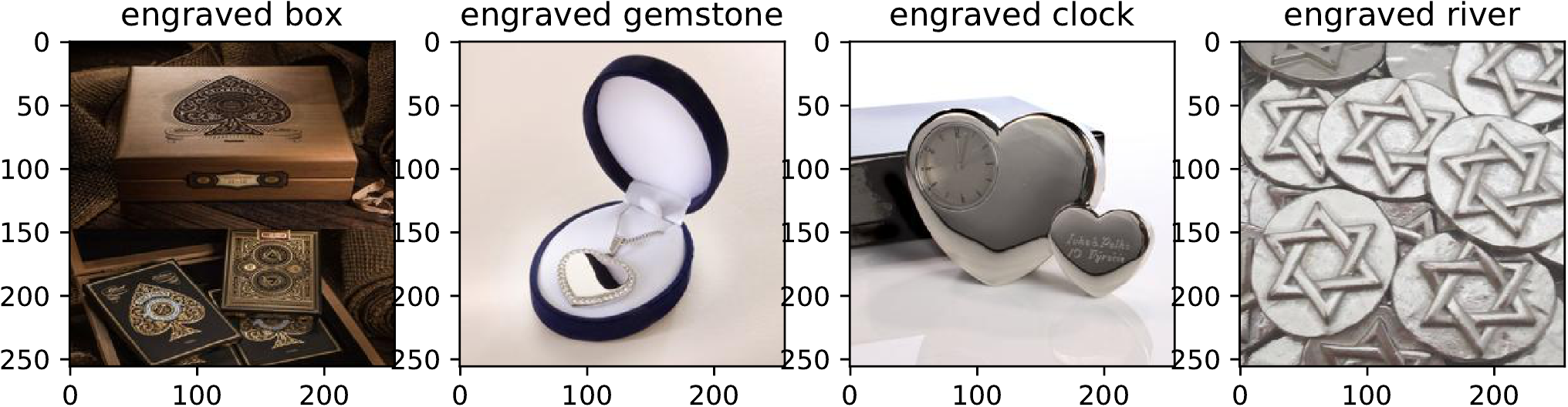} \\~\\
	\includegraphics[width=0.95 \linewidth]{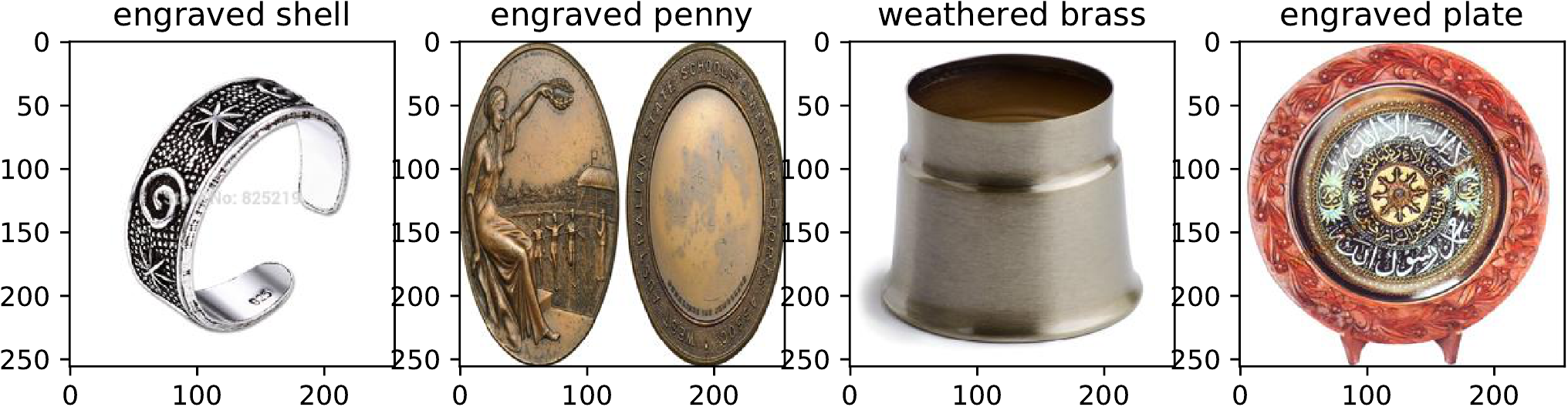} \medskip
	\hrule
	\medskip
	\includegraphics[width=0.95\linewidth]{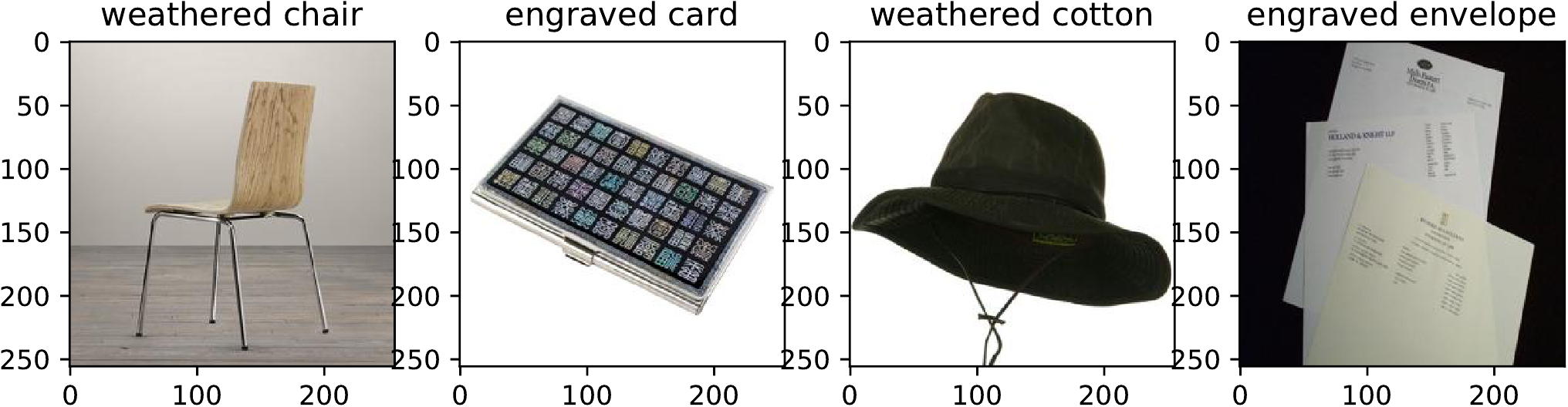} \\~\\
	\includegraphics[width=0.95 \linewidth]{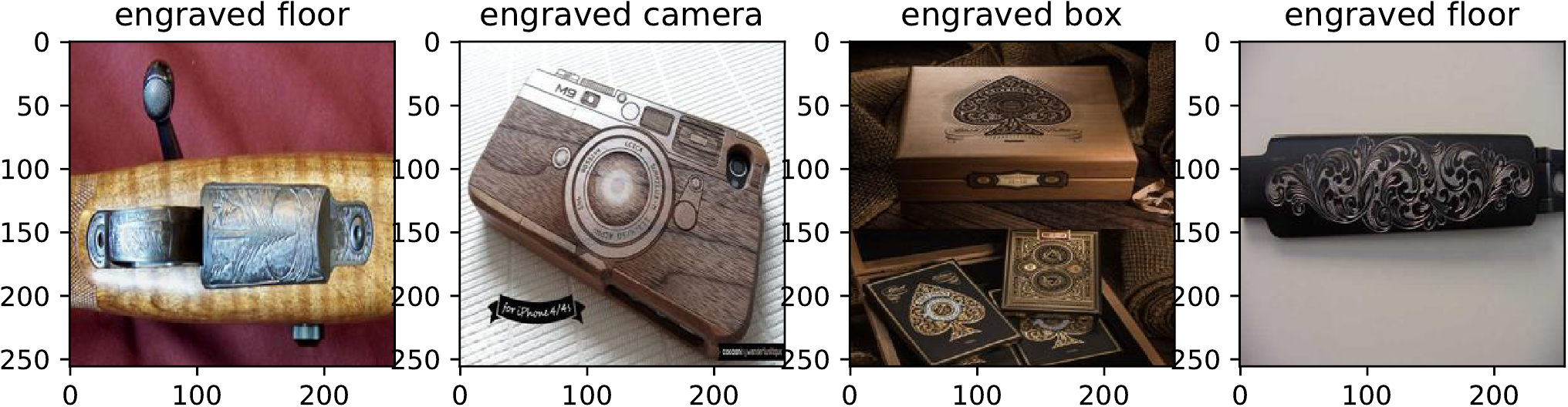}
	\caption{Best matched images obtained through  GW transportation plan, and our SERW distance. The first block of images  correspond to the class {\it Engraved Copper} and the second one to {\it Engraved Metal}. Within each block, the top row shows the results of GW and the bottom row illustrates the matching proposed by SERW.}
	\label{fig:matched_images}
\end{figure}

Figure~\ref{fig:gwtextimage} presents %
coupling matrices obtained
using GW and our SERW. Since in both cases, the %
distance  
is not approximated by the Sinkhorn algorithm, the obtained matching is not smooth. Our results show that both GW and SERW distances are able to retrieve the $3$ classes of adjectives and matches
appropriate images with the relevant labels. 
Figure~\ref{fig:matched_images} illustrates the best matched images by GW and SERW (according to the transportation map) to the texts {\it Engraved Copper} and {\it Engraved Metal}. We can remark that in both cases GW and SERW do not suggest the same images. However, the retrieved images are meaningful according to the text queries. We shall notice that the embeddings used by SERW  do not distort the discriminative information, leading to interesting matched images as shown by the last row of Figure \ref{fig:matched_images}.

\section{Conclusion} %
\label{sec:conclusion}

In this paper we introduced the SERW distance for distribution alignment lying in different mm-spaces. It is based on metric measure embedding of the original mm-spaces into a common Euclidean space and computes an optimal transport on the (low-distorted) embedded distributions. We prove that SERW defines a proper distance behaving like GW distance and we further show a cost relation between SERW and GW. 
Some of numerical experiments are tailored using a fully connected neural network to learn the maximization problem defining SERW, while other ones are conducted with fixed embeddings. 
In particular, SERW can be viewed as an embedding-dependent alignment for distributions coming from different mm-spaces, that its quality is strongly dependent on the given embeddings.

\section*{Acknowledgments}
The works of Maxime Bérar, Gilles Gasso and Alain Rakotomamonjy have been supported by the OATMIL ANR-17-CE23-0012 Project of the French National Research Agency (ANR).
\newpage

\onecolumn
\appendix 

\section{Proofs} %
\label{sec:proofs}

In the proofs, we frequently use the two  following lemmas. Lemma~\ref{lem:integration_pushs} writes an integration result using push-forward measures; it relates integrals with respect to a measure $\eta$ and its push-forward under a measurable map $f: X \rightarrow Y.$
Lemma~\ref{lem:image_admis_couplings} proves that the admissible set of couplings between the embedded measures are exactly the embedded of the admissible couplings between the original measures. 

\begin{lemma}
\label{lem:integration_pushs}
Let $f: S \rightarrow T$ be a measurable mapping, let $\eta$ be a measurable  measure on $S$, and let $g$ be  a measurable function on $T$. Then $\int_T g \diff f_{\#}\eta= \int_S (g\circ f) \diff\eta$.
\end{lemma}

\begin{lemma}
\label{lem:image_admis_couplings}
For all $\phi \in \cF_d(X)$, $\psi \in \cF_d(Y), r\in\cR_d$,  and $\mu \in \mathscr{P}(X), \nu \in \mathscr{P}(Y)$, one has 
\begin{equation*}
\Pi(\phi_{\#}\mu,(r\circ\psi)_{\#}\nu) = \{(\phi \otimes (r\circ\psi))_\# \pi \textrm{ s.t. } \pi \in\Pi(\mu, \nu)\}
\end{equation*}
where $\phi \otimes (r\circ\psi): X \times Y \rightarrow X \times Y$ such that  $(\phi \otimes (r\circ\psi))(x,y) = (\phi(x), r(\psi(y)))$ for all $x,y \in X \times Y.$
\end{lemma}

{\it Proof of Lemma~\ref{lem:image_admis_couplings}.} See \cite{patycuturi2019}.

\subsection{Proof of Proposition~\ref{lem:compact-embedding-sets}} %
\label{sub:_lem_compact_embedding_sets_}

To let it more readable, the proof is divided into 5 steps summarized as follows: first step is for metric property of $\cF_d(X)$; second one shows completeness of $\cF_d(X)$; third establishes the totally boundedness of $\cF_d(X)$, namely that one can recover this space using balls centred on a { finite} number of embedding points; the last is a conclusion using Arzela-Ascoli's Theorem for characterizing compactness of subsets of functional continuous space. %

Since the arguments of the proof are similar for the two spaces, we only focus on proving the topological property of $\cF_d(X)$. 
Let us refresh the memories by some results in topology: we denote $\mathcal{C}(X, \R^d)$ the set of all  continuous mappings of $X$ into $(\R^d , \norm{\cdot})$ and recall the notions of totally boundedness in order to characterize the compactness of $(\cF_d(X), \Gamma_X)$. The material here is taken from~\cite{kubrusly2001elements} and~\cite{o2006metric}.

\begin{definition}
\label{definitions-topo}
{\it i) (Totally bounded)} Let $A$ be a subset of a metric space $(S, d_S).$ A subset $A_\varepsilon$ of A is an $\varepsilon$-net for $A$ if for every point $s$ of $A$ there exists a point $t$ in $A_\varepsilon$ such that $d(s, t) < \varepsilon.$ A subset $A$ of $S $ is {\it totally bounded (precompact)} in $(S, d_S)$ if for every real number $\varepsilon > 0$ there exists a finite $\varepsilon$-net for $A$.\\
{\it ii) (Pointwise totally bounded)} A subset $\mathcal{S}$ of $\mathcal{C}((S, d_S) ,(T, d_T))$ is {\it pointwise totally bounded} if for each $s$ in $S$ the set $\mathcal{S}(s) = \{f(s)\in T: f \in S\}$ is totally bounded in $T.$\\
{\it iii) (Equicontinuous)} A subset $\mathscr{S}$ of $\mathcal{C}(S,T)$ is {\it equicontinuous} at a point $s_0 \in S$ if for each $\varepsilon>0$ there exists a $\delta > 0$ such that $d_T(f(s),f(s_0)) < \varepsilon$ whenever $d_S(s, s_0) < \delta$ for every $f \in \mathscr{S}$
\end{definition}

\begin{proposition}
\label{propsotion:compact-set}
If $S$ is a metric space, then $S$ is compact if and only if $S$ is complete and totally bounded.
\end{proposition}

Let $\mathcal{C}((S, d_S), (T, d_T))$ consisting of all  continuous bounded mappings of $S$ into $(T, d_T),$ endowed with the supremum metric $d_\infty(f,g) = \sup_{s \in S} d_T(f(s),g(s))$. Proving the totally boundedness of some topological spaces may need more technical tricks. Fortunately, in our case we use Arzelà–Ascoli Theorem that  gives compactness criteria for subspaces of $\mathcal{C}((S, d_S), (T, d_T))$ in terms of pointwise totally bounded and equicontinuous, namely they are a necessary and sufficient condition to guarantee that the totally boundedness of a subset $\mathcal{S}$ in $(\mathcal{C}(S, T), d_\infty)$.
\begin{theorem}
\label{theorem-arzelaascoli}
(Arzelà–Ascoli Theorem)
	If $S$ is compact, then a subset of the metric space $\mathcal{C}((S, d_S), (T, d_T))$ is totally bounded if and only if it is pointwise totally bounded and equicontinuous.
\end{theorem}

The proof is devided on 5 Steps: 

 {$\bullet$ \bf Step 1. $(\cF_d(X), \Gamma_X)$ is a metric space.} It is clear that for all $\phi, \phi' \in \cF_d(X)$, $\Gamma_X(\phi, \phi') \geq 0$ (nonegativeness) and $\Gamma_X(\phi, \phi') =0$ if and only if $\phi = \phi'$. To verify the triangle inequality, we proceed as follows. Take and arbitrary $x\in X$ and note that, if $\phi, \phi',$ and $\phi''$ are embeddings in $\cF_d(X)$ then by triangle inequality in the Euclidean space $\R^d$.
\begin{equation*}
\norm{\phi(x) - \phi'(x)} \leq \norm{\phi(x) - \phi''(x)} + \norm{\phi''(x) - \phi'(x)} \leq \Gamma_X(\phi, \phi'') + \Gamma_X(\phi'', \phi'),
\end{equation*}
hence $\Gamma_X(\phi, \phi')\leq \Gamma_X(\phi, \phi'') + \Gamma_X(\phi'', \phi')$, and therefore $(\cF_d(X), \Gamma_X)$ is a metric space. 

{\bf $\bullet$ Step 2. $\cF_d(X) \subset \mathcal{C}(X, \R^d).$} First recall that for each $\phi \in \cF_d(X)$ is a $\tau_\phi$-embedding then it is Lipshitizian mapping. It is readily verified that every Lipshitizian mapping is uniformly continuous, that is for each real number $\varepsilon > 0$ there exists a real number $\delta > 0$ such that $d_X(x, x') < \delta$ implies $\norm{\phi(x) - \phi(x')} < \varepsilon$ for all $x, x' \in X$. So it is sufficient to take $\delta = \frac{\varepsilon}{\tau_\phi}.$ 

{\bf $\bullet$ Step 3.  $(\cF_d(X), \Gamma_X)$ is complete.}  The proof of this step is classic in the topology literature of the continuous space endowed with the supremum metric. 
For the sake of completeness, we adapt it in our case. 
Let $\{\phi_k\}_{k \leq 1}$ be a Cauchy sequence in $(\cF_d(X), \Gamma_X).$ Thus $\{\phi_k(x)\}_{k \leq 1}$ is a Cauchy sequence in $(\R^d, \norm{\cdot})$ for every $x \in X.$ This can be as follows: $\norm{\phi_k(x) - \phi_{k'}}(x)) \leq \sup_{x \in X} \norm{\phi_k(x) - \phi_{k'}}(x)) = \Gamma_X(\phi, \phi')$ for each pair of integers $k$, $k$ and every $x \in X$, and hence $\{\phi_k(x)\}_{k \leq 1}$ converges in $\R^d$ for every $x \in X$ (since $\R^d$ is complete). Let $\phi(x) = \lim_{k \rightarrow \infty} \phi_k(x)$ for each $x\in X$ (i.e., $\phi_k(x) \rightarrow \phi(x))$ in $\R^d$, which defines a a mapping $\phi$ of $X$ into $\R^d$. We shall show that $\phi \in \cF_d(X)$ and that $\{\phi_k\}$ converges to $\phi$ in $\cF_d(X)$, thus proving that $(\cF_d(X), \Gamma_X)$ is complete. Note that for any integer $n$ and every pair of points $x,x'$ in $\cF_d(X)$,
we have $\norm{\phi(x) - \phi(x')} \leq \norm{\phi(x) - \phi_k(x)} + \norm{\phi_k(x) - \phi_k(x')} + \norm{\phi_k(x') - \phi(x')}$ by the triangle inequality.
Now take an arbitrary real number $\varepsilon > 0$. Since $\{\phi_k(x)\}_{k}$ is a Cauchy sequence in  $(\cF_d(X), \Gamma_X)$, it follows that there exists a positive integer $k_\epsilon \in \mathbb{N}$ such that $\Gamma(\phi_{k}, \phi_{k'}) < \varepsilon$, 
and hence $\norm{\phi_k(x) - \phi_{k'}(x)} < \varepsilon$ for all $x \in X$, whenever $k, k' \geq k_\varepsilon$. Moreover, since $\phi_k(x) \rightarrow \phi(x)$ in $\R^d$ for every $x\in X$, and the Euclidean distance is a  continuous function from the metric space $\R^d$ to the metric space $\R$ for each $y \in \R^d$, 
it also follows that $\norm{\phi(x) - \phi_k(x)} = \norm{\lim_{k' \rightarrow \infty}\phi_{k'}(x) - \phi_k(x)}$ for each positive integer $k$ and every $x \in X.$ Thus $\norm{\phi(x) - \phi_{k}(x)} \leq \varepsilon$ for all $x \in X$ whenever $k \geq k_\varepsilon$. Furthermore, since each $\phi_{k}$ lies in $(\cF_d(X), \Gamma_X)$, it follows that there exists a real number $\gamma( k_\varepsilon)$ such that sup $d(\phi_{k_\varepsilon}(s),\phi_{k_\varepsilon}(x')) \leq \gamma( k_\varepsilon),$ $x, x' \in X.$
Therefore, for any $\varepsilon > 0$ there exists a positive integer ${k_\varepsilon}$ such that
$\norm{\phi(x) - \phi(x')} \leq 2\varepsilon + \gamma(k_\varepsilon)$ for all $x, x' \in X$ so that $\phi \in (\cF_d(X), \Gamma_X)$, and $\Gamma_X(\phi, \phi') = \sup_{x \in X} \norm{\phi(x) - \phi'(x)} \leq \varepsilon, x\in X$ whenever $k \geq k_\varepsilon,$ so that $\phi_k$ converges to $\phi$ in $(\cF_d(X), \Gamma_X)$.

{\bf $\bullet$ Step 4.  $\cF_d(X)$ is pointwise totally bounded and equicontinuous.} From $(iii)$ in Definition~\ref{definitions-topo} and the details in {\bf Step 3}, $\cF_d(X)$ is readily equicontinous. 
Next we shall prove that the subset $\{\hat x\}=\{\phi(x) \in \R^d: \phi \in \cF_d(X)\}$ is totally bounded in $\R^d$. To proceed we use another result characterizing totally boundness that reads as: 
\begin{equation*}
\text{\it $(S, d_S)$ is totally bounded metric space if and only if every sequence in $S$ has a Cauchy subsequence.}
\end{equation*}
Since for any $\phi \in \cF_d(X)$ is Lipshitizian then it is uniformly continuous as explained above. Furthermore uniformly continuous functions have some very nice conserving properties. 
They map totally bounded sets onto totally bounded sets and Cauchy sequences onto Cauchy sequences. 

Now suppose that Suppose $\{y_l\}_{l \geq 1}$ is any sequence in $\{\hat x\}\subset \phi(X)$ . For each $l \in \mathbb{N}$, the subset $X \cap \phi^{-1}(\{y_l\}) \subset X$ is non empty for each $l \in\mathbb{N}$~(Axiom of Countable Choice see~\cite{o2006metric}). Then $\phi({x_l}) = y_l$ for each $l \in \mathbb N$. By the Cauchy criterion for total boundedness of $X$, the sequence  $\{x_l\}_{l}$ has a Cauchy subsequence $\{x_l\}_{l_j}$. Then, by what we have just proved, $\{\phi(x_l)\}_{l_j}=\{y_l\}_{l_j}$ is a Cauchy subsequence of $\{y_l\}.$ Since  $\{y_l\}$ is an arbitrary sequence in $\{\hat x\}$, $\{\hat x\}$ satisfies the Cauchy criterion for total boundedness and so is totally bounded.

{\bf $\bullet$ Step 5.  $\cF_d(X)$ is compact.} Using Arzela-Ascoli Thereom~\ref{theorem-arzelaascoli} and {\bf Step 2} we conclude that  $\cF_d(X)$ is totally bounded. Together with {\bf Step 3} $\cF_d(X)$ is compact.

\subsection{Proof of Lemma~\ref{lem:second_def_of_ERW}} %
\label{ssub:proof_of_lemma_lem:second_def_of_erw}

Notice that for $\mu \in \mathscr{P}_p(X)$, $\nu \in \mathscr{P}_p(Y)$, and $(\phi, \psi) \in \cF_d(X) \times \cF_d(Y), r\in \cR_d$ one has $W_2^2\big(\frac 1{\sqrt{2}}\phi_{\#}\mu,\frac 1{\sqrt{2}}(r\circ\psi)_{\#}\nu\big) < \infty.$
It can be seen easily using the facts that 
\begin{equation*}
\int_{\R^d}{\norm{u}^2\diff\frac 1{\sqrt{2}}\phi_{\#} \mu(u)} = \frac 12 \int_{X} \norm{\phi(x)}^2\diff\mu(x)\leq \frac{{\tau_{\phi}^2}}{2} M_2(\mu)
\end{equation*}
and 
\begin{equation*}
\int_{\R^d}{\norm{v}^2\diff\frac 1{\sqrt{2}}(r\circ\psi)_{\#} \nu(u)} = \frac 12 \int_{Y} \norm{r(\psi(y))}^2\diff\nu(y) = \frac 12 \int_{Y} \norm{\psi(y)}^2\diff\nu(y) \leq \frac{{\tau_{\psi}^2}}{2} M_2(\nu)
\end{equation*}
where $M_2(\mu)=\int_{X} \norm{x}_X^2\diff\mu(x)< \infty$ and $M_2(\nu)=\int_{Y} \norm{y}_Y^2\diff\mu(y)< \infty,$ by Assumption~\ref{assumptiom-embeddings}.
Now, thanks to Lemmas~\ref{lem:integration_pushs} and~\ref{lem:image_admis_couplings}, we have 
\begin{align*}
\mathcal{E}^2_d(\mu, \nu)
&= \inf_{r\in \cR_d} \sup_{\phi \in \cF_d(X), \psi \in \cF_d(Y)} \inf_{\gamma \in \Pi(\frac 1{\sqrt{2}}\phi_{\#}\mu,\frac 1{\sqrt{2}}(r\circ \psi)_{\#}\nu)} \int_{X\times Y} \norm{u-v}^2 \diff\gamma(u,v)\\
&=\inf_{r\in \cR_d}\sup_{\phi \in \cF_d(X), \psi \in \cF_d(Y)} \inf_{\pi \in \Pi(\mu,\nu)} \int_{X\times X} \norm{u-v}^2 \diff\big(\frac 1{\sqrt{2}}\phi\otimes \frac 1{\sqrt{2}}(r\circ\psi)\big)_{\#}\pi(u,v)\\
&=\inf_{r\in \cR_d} \sup_{\phi \in \cF_d(X), \psi \in \cF_d(Y)} \inf_{\pi\in \Pi(\mu,\nu)} \frac 12 \int_{X\times Y} \norm{\phi(x)-r(\psi(y))}^2 \diff\pi(x,y)\\
&=\inf_{r\in \cR_d} \sup_{\phi \in \cF_d(X), \psi \in \cF_d(Y)} \frac 12\inf_{\pi\in \Pi(\mu,\nu)} J_{\phi, \psi, r, \pi}(\mu, \nu).
\end{align*}

\subsection{Proof of Lemma~\ref{lem:existence_ERW}} %
\label{sub:proof_of_lemma_lem:existence_erw}

In one hand, for any  fixed $\pi \in \Pi(\mu, \nu)$ the application $h_{\pi}:(\phi, \psi, r) \mapsto \int_{X\times Y}\norm{\phi(x)-r(\psi(y))}^2 \diff\pi(x,y)$ is continuous. To show that, we use the continuity under integral sign Theorem. Indeed,
\begin{itemize}
	\item {for } $\pi\text{-almost } (x,y), \text{ the mapping } (\phi, \psi, r) \mapsto \norm{\phi(x)-\psi(y)}^2 \text{ is continuous.}$ To show that fix $\varepsilon > 0$,  and $\phi, \psi, r, \phi_0, \psi_0, r_0 \in \cF_d(X) \times \cF_d(Y) \times \cR_d.$ We endow the product sapce $\cF_d(X) \times \cF_d(Y) \times \cR_d$ by the metric $\Gamma_{X,Y}$ defined as follows: 
	\begin{equation*}
	\Gamma_{X,Y}((\phi, \psi), (\phi', \psi'), (r, r')) = \Gamma_X(\phi, \phi') + \Gamma_Y(r\circ\psi, r'\circ\psi'))
	\end{equation*}

	 We have 
	\begin{align*}
		\big|\norm{\phi(x)-(r\circ\psi)(y)}^2 - \norm{\phi_0(x)-(r_0\circ\psi_0)(y)}^2\big| &\leq \norm{(\phi(x) - \phi_0(x)) - (r(\psi(y)) - r_0(\psi_0(y)))}^2\\
		&\leq 2 \big(\norm{\phi(x) - \phi_0(x)}^2 +  \norm{r(\psi(y)) - r_0(\psi_0(y))}^2\big)\\
		& \leq 2 \big(\Gamma_X^2(\phi, \phi_0) + \Gamma^2_Y(r\circ\psi, r_0\circ\psi_0)\big)\\
		&\leq 2 \Gamma_{X,Y}^2((\phi, \psi), (\phi_0, \psi_0), (r, r_0)).
	\end{align*}
	Letting $\delta_\varepsilon = \sqrt{\varepsilon/2}$, then if $\Gamma_{X,Y}((\phi, \psi), (\phi_0, \psi_0),(r, r_0)) < \delta_\varepsilon$, one has $\big|\norm{\phi(x)-r(\psi(y))}^2 - \norm{\phi_0(x)-r_0(\psi_0(y))}^2\big| < \varepsilon.$
	This yields that $\lim_{(\phi, \psi, r) \rightarrow (\phi_0, \psi_0, r_0)} \norm{\phi(x)-r(\psi(y))}^2 = \norm{\phi_0(x)-r_0(\psi_0(y))}^2.$

	\item for a fixed $(\phi, \psi, r)$ and $(x,y) \in X\times Y,$ we have $\norm{\phi(x)-r(\psi(y))}^2 \leq \norm{\phi(x)}^2 + \norm{r(\psi(y))}^2\leq g(x,y):= {\tau_\phi}^2 \norm{x}^2 + {\tau_\psi}^2 \norm{y}^2$ with $ \int_{X\times Y} g(x,y)d\pi(x,y) = {\tau_\phi}^2\int_{X}\norm{x}^2\diff\mu(x) + {\tau_\psi}^2\int_{Y}\norm{y}^2\diff\nu(x) < \infty.$
\end{itemize}
Therefore, the family $(h_{\pi})_{\pi \in \Pi(\mu, \nu)}$ is continuous then it is upper semicontinuous. 
We know that the pointwise infimum of a family of upper semicontinuous functions is upper semicontinuous (see Lemma 2.41 in~\cite{aliprantis2006infinite}). This entails $\inf_{\pi \in \Pi(\mu, \nu)} h_{\pi}$ is upper semicontinuous.
Since the product of compact sets is a compact set (Tychonoff Theorem), then $\cF_d(X) \times \cF_d(Y)$ is compact, hence $\sup_{\phi \in \cF_d(X), \psi \in \cF_d(Y)} \inf_{\pi \in \Pi(\mu, \nu)} h_{\pi}(\phi, \psi, r)$ attains a maximum value (see Theorem 2.44 in~\cite{aliprantis2006infinite}). 
So, there exits a couple of embeddings $(\phi^*, \psi^*) \in \cF_d(X) \times \cF_d(y)$ and $\pi^* \in \Pi(\mu, \nu)$ such that $\sup_{\phi \in \cF_d(X), \psi \in \cF_d(Y)} \inf_{\pi \in \Pi(\mu, \nu)} h_{\pi}(\phi, \psi, r) = h_{\pi^*}(\phi^*, \psi^*, r)$ for all $r\in \cR_d.$
Finally, it is easy to show that $r\mapsto h_{\pi^*}(\phi^*, \psi^*, r)$ is continuous, hence the infimum over the orthogonal mappings $\cR_d$ (compact) exits.

\subsection{Proof of Lemma~\ref{lem:minimax-ineq}} %
\label{sub:proof_of_lemma_lem:minimax-ineq}

Let us recall first the minimax inequality:
\begin{lemma}
(Minimax inequality)
Let $\Xi: \cU \times \cV \rightarrow \R\cup \{\pm \infty\}$ be a function. Then
\begin{equation*}
\sup\limits_{v \in \cV} \inf\limits_{u \in \cU} \Xi(u, v) \leq \inf_{u \in \cU} \sup_{v \in \cV} \Xi(u,v).
\end{equation*}
\end{lemma}
Using minimax inequality, one has  
\begin{align*}
\mathcal{E}^2_d(\mu, \nu) %
&\leq \frac 12 \inf_{r \in \cR_d} \inf_{\pi\in \Pi(\mu,\nu)} \sup_{\phi \in \cF_d(X), \psi \in \cF_d(Y)} J_{\phi, \psi, r, \pi}(\mu, \nu). %
\end{align*}
Note that for a fixed $\pi \in \Pi(\mu, \nu)$ and $r\in \cR_d$ one has $\sup_{\phi \in \cF_d(X), \psi \in \cF_d(Y)} J_{\phi, \psi, r, \pi}(\mu, \nu)$ exits (continuity of $\pi, r \mapsto \sup_{\phi \in \cF_d(X), \psi \in \cF_d(Y)} J_{\phi, \psi, r, \pi}(\mu, \nu)$ + compact set as shown in Proof of Lemma~\ref{lem:existence_ERW}). Then
\begin{equation*}
\inf_{r \in \cR_d} \inf_{\pi\in \Pi(\mu,\nu)} \sup_{\phi \in \cF_d(X), \psi \in \cF_d(Y)} J_{\phi, \psi, r, \pi}(\mu, \nu) = \inf_{\pi\in \Pi(\mu,\nu)} \inf_{r \in \cR_d} \sup_{\phi \in \cF_d(X), \psi \in \cF_d(Y)} J_{\phi, \psi, r, \pi}(\mu, \nu).
\end{equation*}
Thus $\mathcal{E}^2_d(\mu, \nu) \leq \mathcal{S}^2_d(\mu, \nu).$

\subsection{Proof of Lemma~\ref{lem:existence_SERW}} %
\label{sub:proof_of_lemma_lem:existence_serw} 

As we proved in~Lemma~\ref{lem:existence_ERW} that for any fixed $\pi \in \Pi(\mu, \nu)$, $h_{\pi}:(\phi, \psi, r) \mapsto \int_{X\times Y}\norm{\phi(x)-r(\psi(y))}^2 \diff\pi(x,y)$ is continuous, then it is lower semicontinous. The pointwise supremum of a family of lower semicontinuous functions is lower semicontinuous (Lemma 2.41 in~\cite{aliprantis2006infinite}) 
Moreover, the pointwise infimum of a compact family of lower semicontinuous functions is lower semicontinuous (here $\cR_d$ is compact) then $\pi \mapsto \inf_{r\in\cR_d} \sup_{\phi \in \cF_d(X), \psi \in \cF_d(Y)}\int_{X\times Y} \norm{\phi(x) - r(\psi(y))}^2 \diff\pi(x,y)$ is lower semicontinuous
Furthermore $\Pi(\mu, \nu)$ is compact set with respect to the topology of narrow convergence~\citep{villani03topics}, then $\inf_{\pi \in \Pi(\mu, \nu)}\inf_{r\in\cR_d}\sup_{\phi \in \cF_d(X), \psi \in \cF_d(Y)}\int_{X\times Y} \norm{\phi(x) - \psi(y)}^2 \diff\pi(x,y)$ exists (see Theorem 2.44 in~\cite{aliprantis2006infinite}).

\subsection{Proof of Proposition~\ref{prop:SERW_vanish}} %
\label{sub:proof_of_lemma_lem:serw_vanish}

$\bullet$ ``$\Rightarrow$'' Suppose that $\mathcal{S}_d(\mu, \nu) =0$ then $\mathcal{E}_d(\mu, \nu) = 0$, that gives the Wasserstein distance $\mathcal{W}_2(\frac{1}{\sqrt{2}}\phi_{\#}\mu, \frac{1}{\sqrt{2}}(r\circ\psi)_{\#}\nu)=0$ and hence $\phi_{\#}\mu = (r\circ\psi)_{\#}\nu$ for any $\phi, \psi \in \cF_d(X) \times \cF_d(Y)$ and $r\in \cR_d.$ 
Then for any $K\subseteq \R^d$ Borel, we have $\mu(\phi^{-1}(K)) = \nu((r\circ\psi)^{-1}(K))$. Recall that $X\subseteq \R^D$ and $Y\subseteq \R^{D'}$, then through the proof lines we regard to $\mu$ and $\nu$ as probability measures on $\R^D$ and $\R^{D'}$, allowing us to use a the following key result of~\cite{cramerwold1936}. 
\begin{theorem}~\citep{cramerwold1936}
Let $\gamma, \beta$ be Borel probability measures on $\R^D$ and agree at every open half-space of $X$. Then $\gamma = \beta$. In other words if, for $\omega \in \mathbb{S}^{D} =\{x\in\R^D: \norm{x} =1\}$ and $\alpha\in\R$ we write $H_{\omega,\alpha} = \{x \in \R^D: \inr{\omega,x} < \alpha\}$ and if $\gamma(H_{\omega,\alpha}) = \beta(H_{\omega,\alpha}),$ for all $\omega \in \mathbb{S}^{D}$ and $\alpha \in \R$ then one has $\gamma=\beta.$
\end{theorem}
The fundamental Cramér-Wold theorem states that a Borel probability measure $\mu$ on $\R^D$ is uniquely determined by the values it gives to halfspaces $H_{\omega, \alpha} = \{ x\in \R^D: \inr{\omega, x} < \alpha\}$ for $\omega \in \mathbb{S}^{D}$ and $\alpha \in \R.$ Equivalently, $\gamma$ is uniquely determined by its one-dimensional projections $(\Delta_\omega)_{\#} \mu$, where $\Delta_\omega$ is the projection $x\in \R^D \mapsto \inr{x, \omega} \in \R$ for $\omega \in \mathbb{S}^{D}$.

Straightforwardly,  we have 
\begin{align*}
\phi^{-1}_{\#}((r\circ\psi)_{\#} \nu)(H_{\omega, \alpha}) &= (r\circ\psi)_{\#} \nu\big((\phi^{-1})^{-1}(H_{\omega, \alpha})\big)\\
&= (r\circ\psi)_{\#} \nu\big(\{u\in X: \phi^{-1}(u) \in H_{\omega, \alpha}\}\big)\\
&= (r\circ\psi)_{\#} \nu\big(\{u\in X: \inr{w,\phi^{-1}(u)} < \alpha\}\big)\\
&= \phi_{\#}\mu\big(\{u\in X: \inr{w,\phi^{-1}(u)} < \alpha\}\big) (\text{ by hypothesis})\\
&= \mu\big(\phi^{-1}\big(\{u\in X: \inr{w,\phi^{-1}(u)} < \alpha\}\big)\big)\\
&= \mu\big(\big\{x\in X: \phi(x) \in \{u\in X: \inr{w,\phi^{-1}(u)} < \alpha\}\big\}\big)\\
&= \mu\big(\{x\in X: \inr{w,\phi^{-1}(\phi(x))} < \alpha\}\big)\\
&= \mu\big(\{x\in X: \inr{w,x} < \alpha\}\big) (\text{ since $\phi$ is one-to-one})\\
&= \mu(H_{\omega, \alpha}).
\end{align*}
Analogously, we prove that $(r\circ\psi)^{-1}_{\#}(\phi_{\#}\mu(H_{\omega, \alpha})) = \nu(H_{\omega, \alpha}).$
Therefore, for all $A \subseteq X$ and $B \subseteq Y$ Borels, we have $\mu(A)=\phi^{-1}_{\#}((r\circ\psi)_{\#}\nu)(A)$ and $\nu(B)= (r\circ\psi)^{-1}_{\#}(\phi_{\#}\mu)(B)$. 

$\bullet$ ``$\Leftarrow$'' Thanks to Lemma~\ref{lem:existence_SERW} in the core of the paper, there exists a couple of embeddings $(\phi^\star, \psi^\star)$ and $r^\star\in \cR_d$ optimum for $\mathcal{S}^2_d(\mu, \nu).$ 
We assume now  that $\nu  = ({(r^\star\circ\psi^\star)}^{-1}  \circ \phi^\star)_{\#} \mu$, then   
\begin{align*}
	\mathcal{S}^2_d(\mu, \nu)
	&= \frac 12\inf_{\pi\in \Pi(\mu,(r^\star\circ\psi^\star)^{-1}_{\#}(\phi^\star_{\#} \mu))}\int_{X\times Y} \norm{\phi^\star(x)-r^\star(\psi^\star(y))}^2 \diff\pi(x,y)\\
	&=\frac 12\inf_{\pi\in \Pi(\mu,\phi^\star_{\#} \mu)}\int_{X\times Y} \norm{\phi^\star(x)-r^\star(\psi^\star(y))}^2 \diff(I \otimes (r^\star\circ\psi^\star)^{-1})_{\#}\pi(x,y)\\
	&=\frac 12\inf_{\pi\in \Pi(\mu, \mu)} \int_{X\times Y} \norm{\phi^\star(x)-r^\star(\psi^\star(y))}^2 \diff(I \otimes \phi^\star)_{\#}\big((I \otimes (r^\star\circ\psi^\star)^{-1})_{\#}\pi(x,y)\big).
\end{align*}
On the other hand, it is clear that $(I \otimes \phi^\star)_{\#}\big((I \otimes (r^\star\circ\psi^\star)^{-1})_{\#}\pi\big)(\cdot)=\big(I \otimes\phi^\star \circ (r^\star\circ\psi^\star)^{-1}\big)_{\#}\pi(\cdot).$
Using the fact that $\phi^\star$ is $\tau_{\phi^\star}$-embedding then we get 
\begin{align*}
\mathcal{S}^2_d(\mu, \nu) &= \frac 12\inf_{\pi\in \Pi(\mu, \mu)}\int_{X\times Y} \norm{\phi^\star(x)-r^\star(\psi^\star(y))}^2 \diff\big(I \otimes\phi^\star \circ (r^\star\circ\psi^\star)^{-1}\big)_{\#}\pi(x,y)\big)\\
&= \frac 12\inf_{\pi\in \Pi(\mu, \mu)}\int_{X\times X} \norm{\phi^\star(x)-\phi^\star(x')}^2 \diff\pi(x,x')\\
&\leq \frac{\tau^2_{\phi^\star}}{2} \inf_{\pi\in \Pi(\mu, \mu)}\int_{X\times X} d^2_X(x, x')\diff\pi(x,x')\\
&\leq \frac{\tau^2_{\phi^\star}}{2} W_2^2(\mu, \mu)\\
&=0.
\end{align*}

\subsection{Proof of Proposition~\ref{prop:ERW_SERW_distances}} %
\label{sub:proof_of_proposition_prop:erw_serw_distances}

Symmetry is clear for both objects. 
In order to prove the triangle inequality, we use a classic lemma known as ``gluing lemma'' that allows to produce a sort of composition of two transport plans, as if they are maps.

\begin{lemma}~\citep{villani03topics}
Let $X, Y, Z$ be three Polish spaces and let $\gamma^1 \in \mathscr{P}(X \times Y)$, $\gamma^2 \in \mathscr{P}(Y \times Z)$, be such that $\Delta^Y_{\#} \gamma^1 = \Delta^Y_{\#}\gamma^2$ where $\Delta^Y$ is the natural projection from $X\times Y$ (or $Y\times Z$) onto $Y$. Then there exists a measure $\gamma \in \mathscr{P}(X \times Y \times Z)$ such that $\Delta^{X\times Y}_{\#} \gamma = \gamma^1$ and $\Delta^{Y\times Z}_{\#} \gamma = \gamma^2$.
\end{lemma}

Let $\eta \in \mathscr{P}_2(Z)$ and $\pi^1 \in \Pi(\mu, \nu)$ and $\pi^2 \in \Pi(\nu, \eta)$. By the gluing lemma we know that there exists $\gamma \in \mathscr{P}_2(X \times Y \times Z)$ such that $\Delta^{X\times Y}_{\#} \gamma = \pi^1$ and $\Delta^{Y\times {Z}}_{\#} \gamma = \pi^2$. Since $\Delta^{X}_{\#} \gamma = \mu$ and $\Delta^{Z}_{\#} \gamma = \eta$, we have $\pi=\Delta^{X \times Z}_{\#} \gamma \in \Pi(\mu, \eta)$.
On the other hand 
\begin{align*}
\int_{X\times Z} &\norm{\phi(x)-\vartheta(\zeta(z))}^2 \diff\pi(x,z)\\
&= \int_{X\times Y \times Z} \norm{\phi(x)-\vartheta(\zeta(z)))}^2 \diff\gamma(x,y,z)\\
&\leq 2\int_{X\times Y \times Z} \big(\norm{(\phi(x)- r(\psi(y))}^2 + \norm{r(\psi(y)) - \vartheta(\zeta(z))}^2 \big)\diff\gamma(x,y,z)\\
&\leq 2\int_{X\times Y \times Z} \norm{\phi(x)- r(\psi(y))}^2\diff\gamma(x,y,z) + 2\int_{X\times Y \times Z} \norm{r({\psi}(y)) - \vartheta(\zeta(z))}^2\diff\gamma(x,y,z)\\
&= 2\int_{X\times Y} \norm{\phi(x)- r(\psi(y))}^2\diff\pi^1(x,y) + 2\int_{Y \times Z} \norm{\tilde{\psi}(y) - \vartheta(\zeta(z))}^2\diff\pi^2(y,z),
\end{align*}
where $\tilde{\psi} = r\circ \psi \in \cF_d(Y)$ ($\norm{r(\psi(y))}_2 = \norm{\psi(y)}_2, \forall y$).
Hence, we end up with the desired result, $	\mathcal{S}^2_d(\mu, \eta) \leq \mathcal{S}^2_d(\mu, \nu) + \mathcal{S}^2_d(\nu, \eta).$

\subsection{Proof of Proposition~\ref{prop:left_equivalence_GW_SERW}} %
\label{sub:proof_of_proposition_prop:equivalence_gw_serw}

As the embedding $\phi$ is Lipschitizian then it is continuous. Since $X$ is compact hence $\phi(X)$ is also compact. Consequently $\text{supp}[\phi_\# \mu] \subset \phi(X)$ is compact (closed subset of a compact). The same observation is fulfilled by $\text{supp}[\psi_\# \nu] \subset \psi(Y)$. Letting $Z = \{\text{supp}[\phi_\# \mu] \cup \text{supp}[(r\circ\psi)_\# \nu]\} \subseteq \R^d$. 
Hence, $(Z, \norm{\cdot})$ is compact metric space and $\phi_\# \mu$ and $(r\circ\psi)_\# \nu$ are 
Borel probability measures on $Z$.
Thanks to Theorem 5 (property (c)) in~\cite{memoli2011GW}, we have that
\begin{equation*}
\mathcal{W}^2_2(\frac{1}{\sqrt{2}}\phi_\# \mu, \frac{1}{\sqrt{2}}(r\circ\psi)_\# \nu)	\geq \mathcal{GW}^2_2(\frac{1}{\sqrt{2}}\phi_\# \mu, \frac{1}{\sqrt{2}}(r\circ\psi)_\# \nu), \text{ for any } \phi \in \cF_d(X), \psi \in \cF_d(Y).
\end{equation*}
So
\begin{equation*}
	\mathcal{E}^2_d(\mu, \nu) \geq \inf_{r\in \cR_d}\sup_{\phi \in \cF_d(X), \psi \in \cF_d(Y)}\mathcal{GW}^2_2(\frac{1}{\sqrt{2}}\phi_\# \mu, \frac{1}{\sqrt{2}}(r\circ\psi)_\# \nu).
\end{equation*}
Together with the minimax inequality we arrive at 
\begin{align*}
&\mathcal{S}^2_d(\mu, \nu)\\
&\geq  \inf_{r\in \cR_d}\sup_{\phi \in \cF_d(X), \psi \in \cF_d(Y)}\mathcal{GW}^2_2(\frac{1}{\sqrt{2}}\phi_\# \mu, \frac{1}{\sqrt{2}}(r\circ\psi)_\# \nu)\\
&\geq  \inf_{r\in \cR_d}\sup_{\phi \in \cF_d(X), \psi \in \cF_d(Y)}\frac 12\inf_{\gamma \in \Pi(\frac{1}{\sqrt{2}} \phi_\# ,\frac{1}{\sqrt{2}} (r\circ\psi)_\#)} \iint_{Z\times Z}(\norm{u - u'}_2- \norm{v - v'}_2)^2 \diff\gamma(u,v) \diff\gamma(u', v')\\
&\geq  \inf_{r\in \cR_d}\sup_{\phi \in \cF_d(X), \psi \in \cF_d(Y)} \frac 14\inf_{\pi\in \Pi(\mu,\nu)}  \iint_{Z\times Z}(\norm{u - u'}_2 - \norm{v - v'}_2)^2 \diff(\phi \otimes (r\circ\psi))_{\#} \pi(u,v) \diff(\phi \otimes (r\circ\psi))_{\#} \pi(u', v')\\
&\geq  \inf_{r\in \cR_d}\sup_{\phi \in \cF_d(X), \psi \in \cF_d(Y)}  \frac 14 \inf_{\pi\in \Pi(\mu,\nu)}\iint_{X\times Y}(\norm{\phi(x) - \phi(x')}_2 - \norm{r(\psi(y)) - r(\psi(y'))}_2)^2 \diff\pi(x,y) \diff\pi(x', y')\\
&= \sup_{\phi \in \cF_d(X), \psi \in \cF_d(Y)}  \frac 14 \inf_{\pi\in \Pi(\mu,\nu)}\iint_{X\times Y}(\norm{\phi(x) - \phi(x')}_2 - \norm{\psi(y) - \psi(y')}_2)^2 \diff\pi(x,y) \diff\pi(x', y')\\
&\geq \sup_{\phi \in \cF_d(X), \psi \in \cF_d(Y)} \frac 14\inf_{\pi\in \Pi(\mu,\nu)} \iint_{X\times Y} \big(d^2_X(x,x') + d^2_Y(y,y')
 - 2 {\tau_\phi\tau_\psi}d_X(x,x')d_Y(y,y')\big) \diff\pi(x,y) \diff\pi(x', y')\\
&\geq \frac 12\mathcal{GW}_2^2(\mu, \nu) +  \frac 12\sup_{\phi \in \cF_d(X), \psi \in \cF_d(Y)}(1- \tau_{\phi}\tau_{\psi})  \inf_{\pi\in \Pi(\mu,\nu)} \iint_{X\times Y} d_X(x,x')d_Y(y,y')\diff\pi(x,y) \diff\pi(x', y').
\end{align*}
Using the fact that $-\sup-x =\inf x$, we get 
\begin{align*}
	\mathcal{GW}_2^2(\mu, \nu) &\leq 2\mathcal{S}^2_d(\mu, \nu) + \inf_{\phi \in \cF_d(X), \psi \in \cF_d(Y)}(\tau_{\phi}\tau_{\psi}- 1) \mathcal{I}(\mu, \nu),
\end{align*}
where $\mathcal{I}_1(\mu, \nu) := \inf_{\pi\in \Pi(\mu,\nu)} \iint_{X\times Y} d_X(x,x')d_Y(y,y')\diff\pi(x,y) \diff\pi(x', y')$. Using Bourgain’s embedding theorem~\cite{bourgain1985}, $\tau_{\phi} \in [1, \bigO(\log n)]$ and $\tau_{\psi} \in [1, \bigO(\log m)]$, then 
\begin{align*}
	\mathcal{GW}_2^2(\mu, \nu) &\leq 2 \mathcal{S}^2_d(\mu, \nu) + \inf_{\tau_{\phi}\in \mathcal{D}_{\text{emb}}(X), \tau_{\psi} \in \mathcal{D}_{\text{emb}}(Y)}(\tau_{\phi}\tau_{\psi}- 1) \mathcal{I}_1(\mu, \nu).
\end{align*}
In another hand, we have 
\begin{align*}
	\mathcal{I}_1(\mu, \nu) &= \inf_{\pi\in \Pi(\mu,\nu)} \iint_{X\times Y} d_X(x,x')d_Y(y,y')\diff\pi(x,y) \diff\pi(x', y')\\
	&\leq \iint_{X\times Y} d_X(x,x')d_Y(y,y')\diff\mu(x) \diff\nu(y)\diff\mu(x') \diff\nu(y')\\
	&\leq \int_{X\times X} d_X(x,x')\diff\mu(x)\diff\mu(x')  \int_{Y\times Y}d_Y(y,y')\diff\nu(y)\diff\nu(y')\\
	&\leq 4 \Big(\int_{X}d_X(x,0)\diff\mu(x) + \int_{Y}d_Y(y,0)\diff\nu(y)\Big)\\
	&\leq 4\Big(\int_{X}\norm{x}_X\diff\mu(x) + \int_{Y}\norm{y}_Y\diff\nu(y)\Big)\\
	&\leq 4 (M_1(\mu) + M_1(\nu)),
\end{align*}
$M_1(\mu) = \int_{X} \norm{x}_X \diff \mu(x)<\infty$. Hence,
\begin{align*}
	\frac 12\mathcal{GW}_2^2(\mu, \nu) &\leq \mathcal{S}^2_d(\mu, \nu) + 2\inf_{\tau_{\phi}\in \mathcal{D}_{\text{emb}}(X), \tau_{\psi} \in \mathcal{D}_{\text{emb}}(Y)}(\tau_{\phi}\tau_{\psi}- 1)(M_1(\mu) + M_1(\nu)).
\end{align*}

\subsection{Proof of Proposition~\ref{prop:right_equivalence_GW_SERW}} %
\label{sub:proof_of_proposition_prop:right_equivalence_gw_serw} 

The proof of this proposition is based on a lower bound for the Gromov-Wasserstein distance (Proposition 6.1 in~\cite{memoli2011GW}):
\begin{equation*}
\mathcal{GW}_2^2 (\mu, \nu) \geq \text{FLB}_2^2(\mu, \nu) := \frac 12 \inf_{\pi \in \Pi(\mu, \nu)} \int_{X\times Y} |s_{X, 2}(x) - s_{Y, 2}(y)|^2 \diff \pi(x, y),
\end{equation*}
where $s_{X,2}: X \rightarrow \R_+$, $s_{X,2}(x') = \Big(\int_{X} d^2_X(x, x')d\mu(x')\Big)^{1/2}$ defines an eccentricity function. Note that $\text{FLB}_2^2$ leads to a mass transportation problem for the cost  $c(x, y) := |s_{X,2}(x) - s_{Y,2}(y|^2.$ 

Now, for any $x, y \in X \times Y$, and $\phi, \psi \in \cF_d(X) \times \cF_d(Y), r\in\cR_d$ we have (by triangle inequality)
\begin{align*}
	\norm{&\phi(x) - r(\psi(y))}_2^2\\
&= \int_{X \times Y} \norm{\phi(x) - r(\psi(y))}_2^2 \diff\mu(x')\diff\nu(y')\\
&\leq 4 \int_{X} \norm{\phi(x) - \phi(x')}_2^2\diff\mu(x') + 4 \int_{Y} \norm{r(\psi(y)) - r(\psi(y'))}_2^2\diff\nu(y')\\
& \qquad + 2 \int_{X \times Y} \norm{\phi(x') - r(\psi(y'))}_2^2\diff\mu(x')\diff\nu(y')\\
&\leq 4 \tau_\phi^2\int_{X} d^2_X(x, x')\diff\mu(x') + 4\tau_\psi^2 \int_{Y} d^2_Y(y, y')\diff\nu(y') + 2 \int_{X \times Y} \norm{\phi(x') - r(\psi(y'))}_2^2\diff\mu(x')\diff\nu(y')\\
&\leq 4 (\tau_\phi^2+\tau_\psi^2) \Big(\int_{X} d^2_X(x, x')\diff\mu(x') + \int_{Y} d^2_Y(y, y')\diff\nu(y')\\
&\qquad - 2 \Big(\int_{X} d^2_X(x, x')\diff\mu(x') \Big)^{1/2} \Big(\int_{Y} d^2_Y(y, y')\diff\nu(y') \Big)^{1/2}\Big)\\
& \qquad + (\tau_\phi^2+\tau_\psi^2)\Big(\int_{X} d^2_X(x, x')\diff\mu(x') \Big)^{1/2} \Big(\int_{Y} d_Y(y, y')^2\diff\nu(y') \Big)^{1/2}\\
&\qquad + 2 \int_{X \times Y} \norm{\phi(x') - r(\psi(y'))}_2^2\diff\mu(x')\diff\nu(y')\\
&\leq 4 (\tau_\phi^2+\tau_\psi^2)\big|s_{X,2}(x) - s_{Y,2}(y)|^2 + 8(\tau_\phi^2+\tau_\psi^2)\sqrt{ \mathcal{I}_{2, x,y}(\mu, \nu)}\\
&\qquad  + 2 \int_{X \times Y} \norm{\phi(x') - r(\psi(y'))}_2^2\diff\mu(x')\diff\nu(y'),
\end{align*}
where 
\begin{align*}
	 \mathcal{I}_{2, x,y}(\mu, \nu) := \Big(\int_{X} d^2_X(x, x')\diff\mu(x')\Big) \Big(\int_{X} Yd^2_Y(y, y')\diff\nu(y')\Big).
\end{align*}
We observe that 
\begin{align*}
\mathcal{I}_{2, x,y}(\mu, \nu) \leq 4(M_2(\mu) + d^2_X(x, 0))(M_2(\nu) + d^2_Y(y, 0)).
\end{align*}
Moreover, 
\begin{align*}
	\int_{X \times Y} \norm{\phi(x') - r(\psi(y'))}^2\diff\mu(x')\diff\nu(y') \leq 2(\tau_\phi^2+ \tau_\psi^2) (M_2(\mu) + M_2(\nu)).
\end{align*}
Therefore, for any $\pi \in \Pi(\mu, \nu)$
\begin{align*}
\int_{X\times Y}&\norm{\phi(x) - r(\psi(y))}_2^2\diff\pi(x,y)\\ &\leq 2(\tau_\phi^2+\tau_\psi^2)\int_{X, Y}\big|s_{X,2}(x) - s_{Y,2}(y)|^2d\pi(x,y)\\
&\quad + 8(\tau_\phi^2+\tau_\psi^2)\int_{X\times Y}\sqrt{4(M_2(\mu) + d^2_X(x, 0))(M_2(\nu) + d^2_Y(y, 0))} \diff\pi(x,y)\\
&\quad + 2(\tau_\phi^2+\tau_\psi^2) (M_2(\mu) + M_2(\nu))\\
&\leq 	4 (\tau_\phi^2+\tau_\psi^2)\int_{X, Y}\big|s_{X,2}(x) - s_{Y,2}(y)|^2d\pi(x,y)\\
&\quad + 16(\tau_\phi^2+\tau_\psi^2)\int_{X}\sqrt{(M_2(\mu) + d^2_X(x, 0))}\diff\mu(x)\int_Y\sqrt{(M_2(\nu) + d^2_Y(y, 0))} \diff\nu(y)\\
&\quad + 2(\tau_\phi^2+\tau_\psi^2) (M_2(\mu) + M_2(\nu)).
\end{align*}
Note that 
\begin{align*}
\int_{X}\sqrt{M_2(\mu) + d^2_X(x, 0)}\diff\mu(x) \leq \sqrt{M_2(\mu)} + \int_{X} d_X(x, 0))\diff\mu(x)\leq \sqrt{M_2(\mu)} + \sqrt{M_1(\mu)},
\end{align*}
and 
\begin{align*}
\int_{Y}\sqrt{M_2(\nu) + d^2_Y(y, 0)}\diff\nu(y) \leq \sqrt{M_2(\nu)} + \int_{Y} d_Y(y, 0))\diff\nu(y)\leq \sqrt{M_2(\nu)} + \sqrt{M_1(\nu)}.
\end{align*}
So
\begin{align*}
\int_{X\times Y}\norm{\phi(x) - r(\psi(y))}_2^2\diff\pi(x,y) &\leq 4 (\tau_\phi^2+\tau_\psi^2)\int_{X, Y}\big|s_{X,2}(x) - s_{Y,2}(y)|^2\diff\pi(x,y)\\	
&\quad + 16(\tau_\phi^2+\tau_\psi^2) (\sqrt{M_2(\mu)} + \sqrt{M_1(\mu)}) (\sqrt{M_2(\nu)} + \sqrt{M_1(\nu)})\\
&\quad + 2(\tau_\phi^2+\tau_\psi^2) (M_2(\mu) + M_2(\nu)).
\end{align*}
Finally,
\begin{align*}
\mathcal{S}^2_d(\mu, \nu) \leq2\sup_{\tau_{\phi}\in \mathcal{D}_{\text{emb}}(X), \tau_{\psi} \in \mathcal{D}_{\text{emb}}(Y)}(\tau_\phi^2 +\tau_\psi^2) (\mathcal{GW}^2_2(\mu, \nu) + \Uoverline{{M}}_{\mu,\nu}),
\end{align*} 
where 
\begin{align*}
\Uoverline{M}_{\mu,\nu} &= 8(\sqrt{M_2(\mu)} + \sqrt{M_1(\mu)}) (\sqrt{M_2(\nu)} + \sqrt{M_1(\nu)}) + (M_2(\mu) + M_2(\nu)).
\end{align*}
This finishes the proof.

\subsection{Proof of Lemma~\ref{lem:fix-emb-distance}} %
\label{sub:proof_of_lemma_lem:fix-emb-distance}

Since $\cR_d$ is compact set and the mapping $r\mapsto J_{\phi_f, \psi_f, r, \pi}(\mu, \nu)$ is continuous, then there exists $r_f \in \cR_d$ such that $\inf_{r \in \cR_d} J_{\phi_f, \psi_f, r, \pi}(\mu, \nu) = J_{\phi_f, \psi_f, r_f, \pi}(\mu, \nu)$. Using Lemma~\ref{lem:image_admis_couplings}, we then get 
\begin{align*}
\widetilde{\cS^2_d}(\mu, \nu) &= \frac 12\inf_{\pi\in \Pi(\mu,\nu)} J_{\phi_f, \psi_f, r_f,  \pi}(\mu, \nu)\\
& = \frac 12\inf_{\pi\in \Pi(\mu,\nu)} \int_{X\times Y} \norm{\phi_f(x) - r_f(\psi_f(y))}^2 \diff \pi(x,y)\\
&= \inf_{\gamma \in \Pi(\frac 1{\sqrt{2}}(\phi_f){\#}\mu,\frac 1{\sqrt{2}}(r_f\circ \psi_f)_{\#}\nu)} \int_{\R^d\times \R^d} \norm{u-v}^2 \diff\gamma(u,v)\\
&= \mathcal{W}_2^2(\mu_f, \nu_f),
\end{align*}
where $\mu_f = (\frac 1{\sqrt{2}}\phi_f)_{\#}\mu$ and $\nu_f = (\frac 1{\sqrt{2}}r_f\circ \psi_f)_{\#}\nu$. Therefore, $\widetilde{\cS^2_d}(\mu, \nu)$ is the 2-Wasserstein distance between $\mu_f$ and $\nu_f$. Hence $\widetilde{\cS^2_d}(\mu, \nu) = 0$ if and only if $\mu_f = \nu_f$ that is ${\phi_f}_{\#}\mu = (r_f\circ \psi_f)_{\#}\nu$. On the other hand, one has 
\begin{align*}
\mu = ({\phi_f}^{-1} \circ \phi_f)_{\#} \mu = ({\phi_f}^{-1})_{\#}({\phi_f}_{\#}\mu)
&= ({\phi_f}^{-1})_{\#}((r_f\circ \psi_f)_{\#}\nu) = ({\phi_f}^{-1} \circ(r_f\circ \psi_f))_{\#}\nu.
\end{align*}
The triangle inequality follows the same lines as proof of Proposition~\ref{prop:ERW_SERW_distances}.

\bibliography{biblio}
\bibliographystyle{chicago}

\end{document}